\documentclass[preprint,review,times]{elsarticle}

\biboptions{sort&compress}
\usepackage{amssymb}
\usepackage{url}
\usepackage[colorlinks]{hyperref}
\biboptions{sort&compress}
\usepackage{dsfont}
\usepackage{amsmath}
\usepackage{siunitx}
\DeclareFontFamily{U}{mathx}{\hyphenchar\font45}
\DeclareFontShape{U}{mathx}{m}{n}{<-> mathx10}{}
\DeclareSymbolFont{mathx}{U}{mathx}{m}{n}

\usepackage{graphbox}
\usepackage[export]{adjustbox}
\newlength{\tempdima}
\newcommand{\rowname}[1]{\rotatebox{90}{\makebox[\tempdima][c]{#1}}}
\usepackage{caption}
\usepackage{subcaption}
\usepackage{booktabs}
\usepackage[super]{nth}
\usepackage{dashrule}
\usepackage{nicematrix,tikz,booktabs}
\usepackage{enumitem}
\usepackage{longtable}
\usepackage{setspace}
\usepackage{etoolbox}
\preto\longtable{\par\singlespacing}

\newcommand{\interval}[2]{\left[#1, #2\right]}
\newcommand{\intervaldeg}[2]{\left[#1^\circ, #2^\circ\right]}

\usepackage{footmisc}
\usepackage{multirow}

\makeatletter
\newcommand\extralabel[2]{{\edef\@currentlabel{\@currentlabel#2}\label{#1}}}
\makeatother

\makeatletter
\renewcommand{\fnum@figure}{Fig. \thefigure}
\makeatother

\journal{Computers in Biology and Medicine}

\graphicspath{{images/}}

\begin{document}

\begin{frontmatter}



\title{Resource Efficient Multi-stain Kidney Glomeruli Segmentation via Self-supervision} 



\author[1]{Zeeshan Nisar\texorpdfstring{\corref{cor1}}{}} 
\cortext[cor1]{Corresponding author}
\ead{nisar@unistra.fr}
\author[2]{Friedrich Feuerhake}
\author[1]{Thomas Lampert}
\affiliation[1]{organization={ICube, University of Strasbourg, CNRS (UMR 7357)}, country={France}}
\affiliation[2]{organization={Institute of Pathology, Hannover Medical School}, country={Germany}}

\begin{abstract}
Semantic segmentation under domain shift remains a fundamental challenge in computer vision, particularly when labelled training data is scarce. This challenge is particularly exemplified in histopathology image analysis, where the same tissue structures must be segmented across images captured under different imaging conditions (stains),
each representing a distinct visual domain. Traditional deep learning methods like UNet require extensive labels, which is both costly and time-consuming, particularly when dealing with multiple domains (or stains). To mitigate this, various unsupervised domain adaptation based methods such as UDAGAN have been proposed, which reduce the need for labels by requiring only one (source) stain  to be labelled. Nonetheless, obtaining source stain labels can still be challenging. This article shows that through self-supervised pre-training---including SimCLR, BYOL, and a novel approach, HR-CS-CO---the performance of these segmentation methods (UNet, and UDAGAN) can be retained even with 95\% fewer labels. Notably, with self-supervised pre-training and using only 5\% labels, the performance drops are minimal: $5.9\%$ for UNet and $6.2\%$ for UDAGAN, averaged over all stains, compared to their respective fully supervised counterparts (without pre-training, using 100\% labels). Furthermore, these findings are shown to generalise beyond their training distribution to public benchmark datasets. Implementations and pre-trained models are publicly available \href{https://github.com/zeeshannisar/resource-effecient-multi-stain-kidney-glomeruli-segmentation.git}{online}.
\end{abstract}




\end{frontmatter}

\section{Introduction} \label{sec:intro}
Deep learning has transformed the field of computer vision, achieving remarkable success in tasks such as image recognition, object detection and semantic segmentation \cite{manakitsa2024review}. Among these, \textit{semantic segmentation}---the task of assigning a class label to every pixel in an image---has particularly benefited from the development of deep learning architectures such as U-Net \cite{ronneberger2015u} and its numerous variants. These architectures have shown outstanding performance across a broad spectrum of real-world applications, including satellite imaging, autonomous driving, and medical image analysis. However, this success relies heavily on the availability of large scale, pixel-level annotated datasets. Acquiring such datasets remains a major bottleneck, as annotation requires substantial time, computational resources, and domain-specific expertise.

This limitation becomes even more pronounced in \textit{multi-domain} scenarios, where semantically identical structures must be segmented across images obtained under varying imaging conditions, modalities or acquisition protocols. In such scenarios, models trained on a single domain often generalise poorly to other domains due to \textit{domain shift} \cite{nisar2022towards}. A straightforward, albeit costly solution, to address these limitations involves acquiring sufficient labels for each domain, followed by the subsequent training of distinct deep learning models tailored for each domain. However, this approach proves highly impractical and inefficient, as it undermines the inherent efficiency and potential generalisation capability of deep learning methods \cite{zhou2022domain}. Additionally, creating and training individual models for each domain can be exceedingly complex, since each model has to be trained on labelled data for each domain, which is a costly and time consuming process---particularly in domains where expert knowledge is indispensable and annotation costs are high, such as medical imaging and microscopy.

\begin{figure}[tb] 
    \centering
    \setlength{\tabcolsep}{1.5pt}
    \begin{tabular}{c c c} 
        H\&E & Jones H\&E & PAS
        \\
        \includegraphics[width=0.2\textwidth]{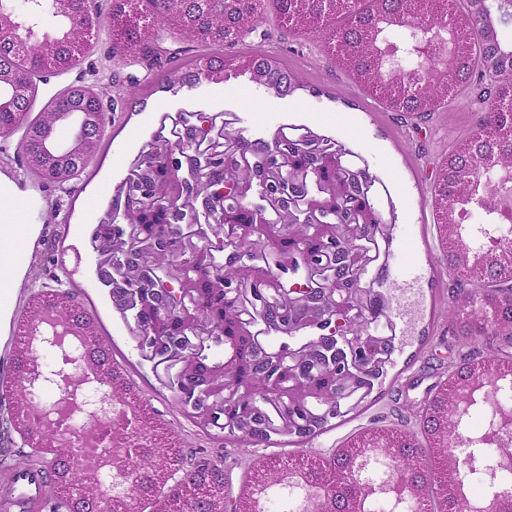} & 
        \includegraphics[width=0.2\textwidth]{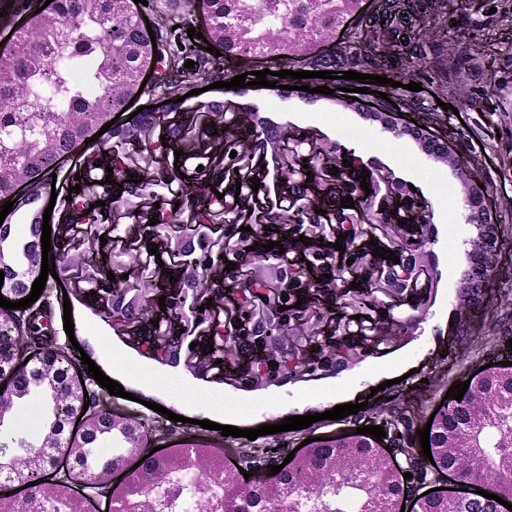} & 
        \includegraphics[width=0.2\textwidth]{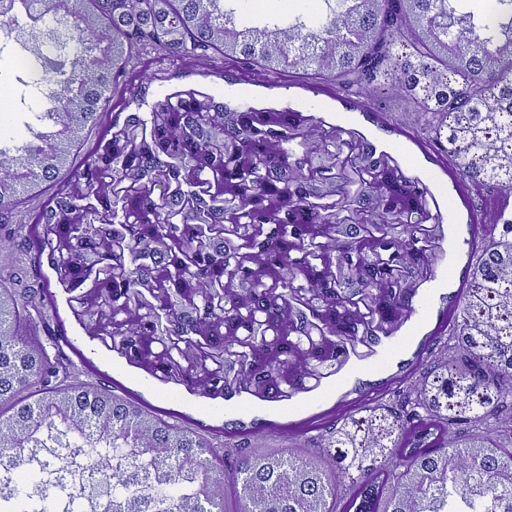} 
        \\ 
        Sirius Red & CD68 & CD34
        \\
        \includegraphics[width=0.2\textwidth,valign=c]{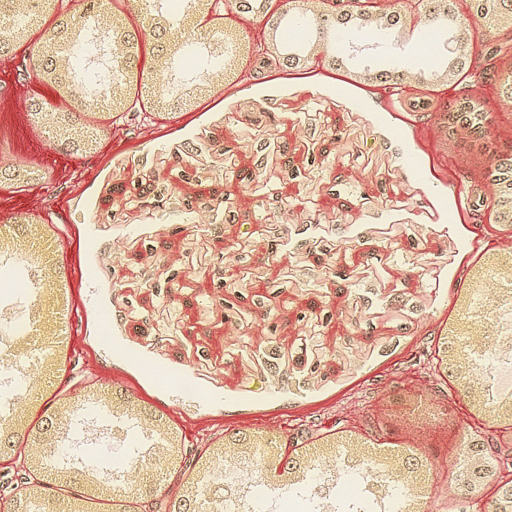} &
        \includegraphics[width=0.2\textwidth,valign=c]{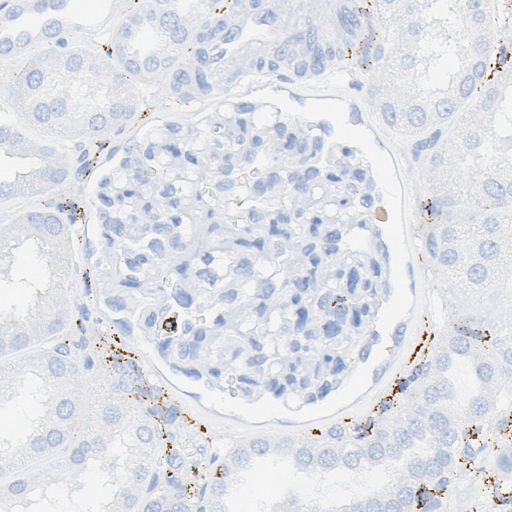} & 
        \includegraphics[width=0.2\textwidth,valign=c]{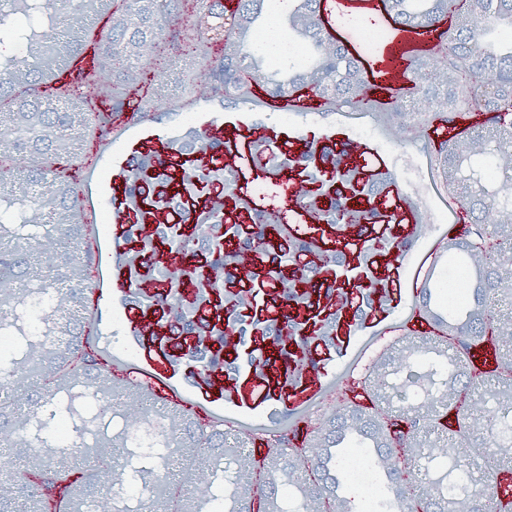}
    \end{tabular} 
	\caption{Different stains used in kidney pathology. Each image represents a glomerulus and each stain provides specific information about the structure of glomerulus.}
	\label{fig:inter_stain_variations}
\end{figure}

A prominent example of these challenges arises in \textit{histopathology image analysis}, where tissue sections are examined microscopically using a variety of staining protocols to highlight different cellular and structural features. Each staining protocol represents a unique imaging domain, producing visually different appearances of the same underlying tissue (see Fig.\ \ref{fig:inter_stain_variations}). Pathologists often analyse different stains to gain a comprehensive understanding of tissue organisation, cellular composition, and disease progression. For instance, in kidney pathology, glomeruli---the key functional units of the kidney---are examined across various stains to assess structural and pathological variations that are essential for accurate diagnosis and disease characterisation. 

Automating this workflow requires the accurate segmentation of glomerular structures across multiple stains, a task for which UNet based architectures are popular \cite{de2018automatic} and still offer state-of-the-art performance \cite{isensee2021nnu, chen2024tinyu, CHEN2024103280}, provided that large amount labels are available for each stain. However, as discussed earlier, labelling medical datasets is complex, can require expert knowledge, and be subject to privacy concerns. As a result, the scarcity of labelled data presents a significant barrier to the scalable and efficient adoption of these deep learning solutions in histopathological image analysis. 

A popular strategy to overcome these challenges is \textit{unsupervised domain adaptation} (UDA), which aims to transfer knowledge from a labelled source domain (stain) to one or more unlabelled target domains (stains) by learning domain-invariant representations. In histopathology, UDA is often achieved through stain transfer techniques, and several stain transfer based multi-stain segmentation methods \cite{gadermayr2019generative, vasiljevic2021towards, bouteldja2022improving, yang2025unsupervised} have been proposed. 
Among them, Unsupervised Domain Augmentation using Generative Adversarial Networks (UDAGAN) \cite{vasiljevic2021towards} is the most prominent approach, which enables the training of a single stain-invariant segmentation model on multiple stains using labels from only the source stain. These methods
have shown promising results among all stains, achieving performance levels comparable to those of fully supervised U-Net models, as will be shown later in this study.
	
While these methods effectively eliminate the reliance on labelled data from target domain(s), they still depend on a substantial amount of labelled data from the source domain, which can be difficult to obtain---especially for newly emerging imaging applications or rare diseases. As a result, the application of existing multi-domain segmentation approaches may prove impractical, creating a significant barrier to their widespread adoption. This exemplifies a broader computer vision challenge: how can we develop robust segmentation models that generalise across multiple domains whilst minimising annotation requirements as much as possible, even for the source domain?

Although expert annotations are often difficult and costly to obtain, the availability of unlabelled datasets has increased substantially in recent years. For instance, in histopathology, the advent of whole slide imaging (WSI) scanners has enabled the routine production of vast repositories of tissue images, presenting an opportunity to alleviate the constraints of scarce annotations through state-of-art representation learning methods. In particular, self-supervised learning (SSL) offers a promising avenue for leveraging unlabelled data to learn representations via pretext tasks \cite{chen2019self}, which are subsequently refined for downstream tasks such as classification and segmentation using a very few labels. 
	
Building on this idea, the primary focus of this study is to demonstrate that unlabelled histopathological data can be effectively used to counteract the effects of a lack of labelled data in segmentation tasks, particularly in the application of kidney glomeruli segmentation across multiple stains. To this end, we evaluate state-of-art SSL methods, such as SimCLR \cite{chen2020simple}, BYOL \cite{grill2020bootstrap}, selected for their general-purpose, domain-agnostic design, and CS-CO \cite{yang2022cs}, chosen for its histopathology oriented formulation. Because CS-CO is stain-specific, we introduce HR-CS-CO, an extension that overcomes its stain dependence to support multi-stain pre-training. These pre-trained models are then fine-tuned for two downstream segmentation tasks: (a) single-stain segmentation with UNet; and (b) multi-stain segmentation with UDAGAN, with our primary emphasis on the latter---which, to the best of our knowledge, has not previously been explored under SSL pre-training.

Notably, this article will show that  when pre-trained with SSL and fine-tuning with only a few labels (5\%), performance comparable to fully (100\%) supervised models can be obtained in both of the above-mentioned tasks. For instance, five UNets fine-tuned with only 5\% labels, each from one of five different stains, achieve an average F$_1$ score of 0.810 across all stains, compared to 0.869 for their fully supervised counterparts (each trained with 100\% labels). Similarly, fine-tuned UDAGAN models achieve average F$_1$ scores of 0.765, across five stains, using only 5\% labels from one (source) stain, whereas the respective baselines trained using 100\% labels achieve 0.827. These findings demonstrate that self-supervised fine-tuned models can approach the performance of fully supervised models, while significantly reducing label requirements by up to 95\%. 

Additionally, our generalisation study reveals that the benefits of pre-trained models extend to diverse datasets and domains. For instance, on the HuBMAP Kaggle competition dataset, fine-tuned UNet models with just 5\% labels achieve F$_1$ score of 0.816, compared to 0.947 for fully supervised baseline and 0.951 for the competition winning model (trained with 100\% labels). Similarly, on the KPIs Challenge dataset, fine-tuned UNets reach an F$_1$ score of 0.795 using only 5\% labels, outperforming the corresponding fully supervised baseline model (0.786) and approaching the top-performing models score of 0.886. These findings further highlight the effectiveness of these pre-trained models in learning robust and transferable features that enable competitive performance on external datasets even under extreme annotation scarcity.   

To achieve this, this study makes the following contributions:
\begin{enumerate}[wide, labelwidth=!, itemindent=!, labelindent=0pt]
    \item Develop a new formulation for CS-CO (a self-supervised pre-text task), called HR-CS-CO, to overcome its stain-specific limitation, thus making it stain independent.
    \item Conduct the first comprehensive analysis of SSL's impact on multi-stain histopathology segmentation, using the use case of kidney glomeruli segmentation to deduce the most appropriate self-supervised approach. 
    \item Reduce the annotation requirement of both single-stain UNet  and multi-stain UDAGAN models by 95\% while maintaining competitive performance. 
    \item Public release of multi-stain kidney pathology pretrained models, enabling the research community to fine-tune downstream segmentation and classification tasks---particularly for glomeruli, with potential extension to other renal structures---on in-house or public datasets using only a few labels.
\end{enumerate}
The rest of the article is organised as follows: Section \ref{sec:ssl} reviews the literature on SSL; Section \ref{sec:methods} provides the architectural details of the aforementioned self-supervised pre-training methods and downstream segmentation tasks; Section \ref{sec:experimental_setup} provides an overview of the dataset and explains the training details; Section \ref{sec:results} evaluates the effectiveness of self-supervised pre-training and provide a detailed comparison to their baseline results; Section \ref{sec:generalisation} examines how well this pre-training generalises across several external datasets; Section \ref{sec:discussions} discusses these results alongside the limitations and potential applications of our study; and finally, Section \ref{sec:conclusion} concludes the article.

\section{Self-Supervised Learning} \label{sec:ssl}
This section presents an overview of SSL methods for visual representation learning, in which designing a pretext task is key. Based on the type of pretext task, existing SSL methods can be categorised into three groups: generative, discriminative, and multi-tasking. For an in-depth review, please see 
\cite{shurrab2022self}.

\subsection{Generative Self-supervised Learning}
This group of pretext tasks learns representations by either reconstructing the original input or by learning to generate samples. Auto-encoders and generative adversarial networks (GANs) are commonly used to achieve these objectives. Several such tasks have been proposed in the field of computer vision and medical imaging, e.g.\ context restoration \cite{chen2019self}, image denoising \cite{prakash2020leveraging}, visual field expansion \cite{boyd2021self} and image inpainting \cite{hu2020self}, etc. However, these approaches are computationally expensive, complex, and may not be necessary when the goal is to learn a simple lower-dimensional data representation \cite{chen2020simple}. Furthermore, they tend to favour low-level features, which are not effective for discriminative downstream tasks \cite{yang2022cs}. To address this, various discriminative SSL methods have been introduced.

\subsection{Discriminative Self-supervised Learning}
Earlier development of these methods focused on context based (or predictive) methods, which showed limited success in medical imaging applications despite significant gains in the computer vision domain \cite{shurrab2022self}. This disparity stems from fundamental differences between natural and medical images, including distinct visual patterns, textures, lighting conditions, and scales. To this end, contrastive learning has emerged as a powerful discriminative approach in both computer vision and medical imaging, particularly in digital histopathology \cite{stacke2021learning, ciga2022self}. It learns representations by comparing pairs of input samples to maximise the similarity between similar (positively-paired) samples and minimising it between dissimilar (negatively-paired) samples. Positive pairs are created by augmenting an input image and negative pairs by selecting different images. As such, the positive pairs preserve global features, which encourages the model to discard irrelevant features and focus on learning representations that are discriminative and robust. Notable approaches to this are Contrastive Predictive Coding \cite{lu2020semi}, A Simple Framework for Contrastive Learning of Visual Representations (SimCLR) \cite{chen2020simple}, and Bootstrap Your Own Latent (BYOL) \cite{grill2020bootstrap} etc.\ 

\subsection{Multi-tasking/Hybrid Self-supervised Learning}
This group is increasing in popularity, in which multiple self-supervised tasks are unified or integrated individually. This can increase the robustness of the representations by reducing bias inherent in individual self-supervised pre-text tasks. Moreover, integrating different approaches to SSL (predictive, generative, and contrastive) improves the model's ability to capture both low-level and high-level features to simultaneously address multiple objectives. For instance, \citet{graham2023one} used multi-tasking to achieve disease classification and segmentation in the same framework. \citet{zhang2021twin} combine predictive and contrastive SSL tasks and \citet{koohbanani2021self} combined multiple predictive tasks. Similarly, \citet{yang2022cs} and \citet{YU2024110327} combined generative and discriminative tasks to extract more robust representations. 

\section{Methods} \label{sec:methods}
\subsection{Self-supervised Pre-training} \label{sec:methods::subsec:ssl}
This study uses three approaches to self-supervised pre-training: SimCLR \cite{chen2020simple}, BYOL \cite{grill2020bootstrap} and CS-CO  \cite{yang2022cs}. SimCLR is selected due to its widespread adoption in histopathology related downstream tasks \cite{stacke2021learning, ciga2022self}. While SimCLR's performance heavily depends on augmentation and the number of negative pairs, demanding huge computing resources \cite{chen2020simple, stacke2021learning}, recent methods have shown that negative-pairs are not essential for contrastive learning \cite{grill2020bootstrap, chen2021exploring, LI2023109364}. One such method is BYOL \cite{grill2020bootstrap}. The original architectures, as proposed by the authors of SimCLR and BYOL, were used in this study, with details provided in \ref{appendix:subsubsec:simclr} and \ref{appendix:subsubsec:byol} 
respectively. Finally, to assess the benefits of hybrid SSL strategies, a novel extension to CS-CO \cite{yang2022cs} is included. This extension (presented in Section \ref{methods:csco}) overcomes its stain specific formulation (it was proposed for the H\&E stain) and is called HR-CS-CO.

\subsubsection{CS-CO} \label{methods:csco}
CS-CO \cite{yang2022cs} is a hybrid SSL method, designed particularly for Haematoxylin and Eosin (H\&E) stained histopathology images. It contains two stages: cross-stain prediction and contrastive learning. The cross-stain prediction, which is a generative task, captures low-level general features, e.g.\ nuclei morphology and tissue texture, that are valuable for histopathology analysis \cite{yang2022cs}. To facilitate this, stain-separation \cite{vahadane2016structure} is applied to H\&E stained images to extract the single-dye channels, Haematoxylin ($H_{ch}$) and Eosin ($E_{ch}$). Afterwards, cross-stain prediction is employed to learn the relationship between $H_{ch}$ and $E_{ch}$, using two separate auto-encoders, H2E and E2H, where H2E predicts $E_{ch}$ from $H_{ch}$, and vice-versa. 

Nevertheless, CS-CO has certain  limitations, restricting its broader applicability. For example, histopathological images often use different staining protocols and reagents to highlight different tissue structures (e.g.\ PAS, Jones H\&E, Sirius Red, CD68, and CD34, as used in this study). The stain separation method integral to CS-CO struggles with ImmunoHistochemical (IHC) stains \cite{vahadane2016structure}. Particularly, it fails to accurately extract the individual $H_{ch}$ and $DAB_{ch}$ (Diaminobenzidine) from CD68. Furthermore, in some cases, histopathological stains contain more than two dyes, e.g.\ Jones H\&E, where CS-CO's stain-separation approach would yield three separate channels---$J_{ch}$ (Jones), $H_{ch}$, and $E_{ch}$---which cannot be handled in CS-CO's architecture. 

\begin{figure}[!tb]
    \centering
    \includegraphics[width=\textwidth]{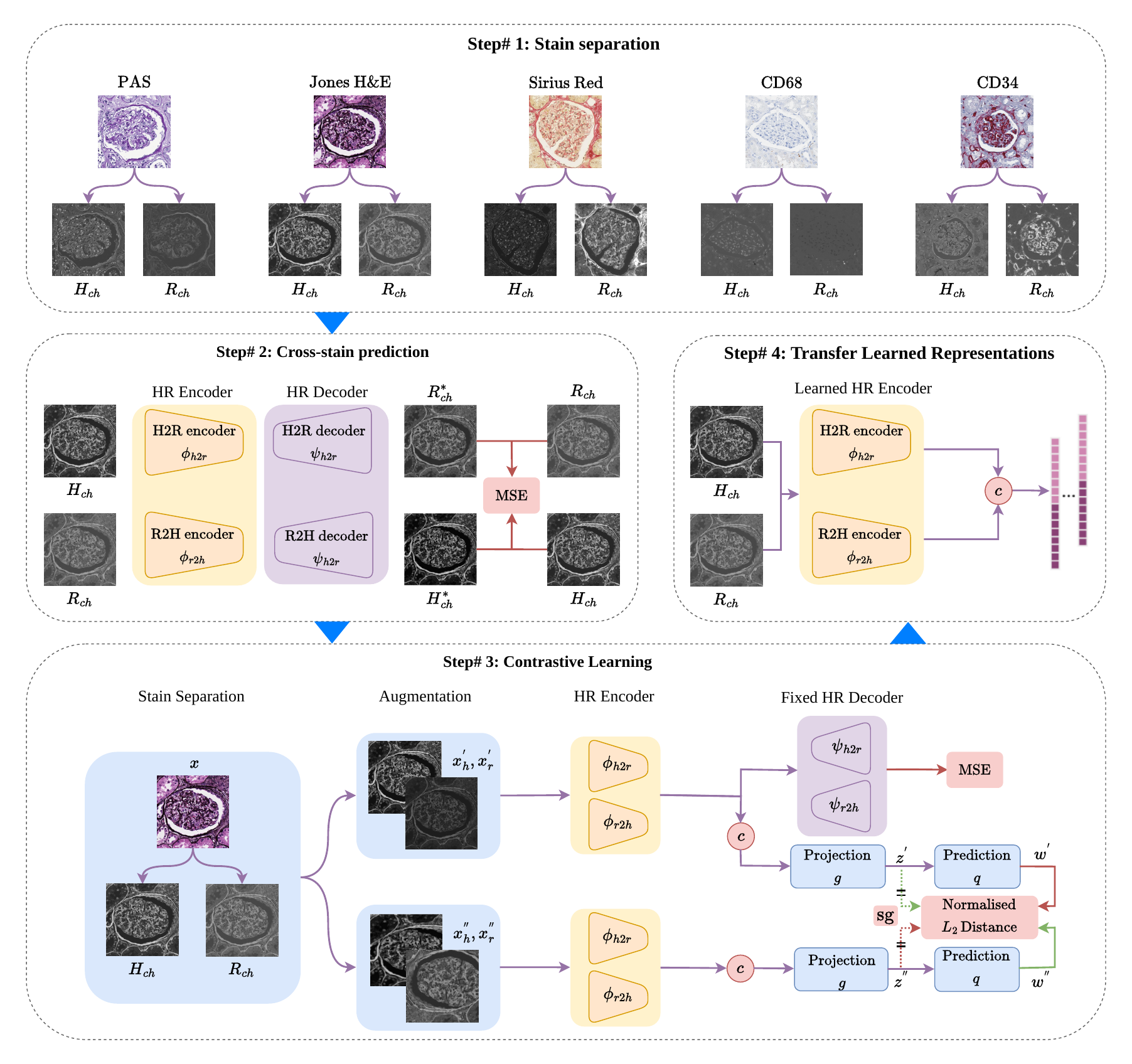}
    \caption{Overview of the proposed HR-CS-CO architecture. In Step\# 1, stain-separation is applied to separate the $H_{ch}$ and $R_{ch}$ from each each stain. In Step\# 2, the cross-stain prediction is employed as a generative task, learning to predict $H_{ch}$ from $R_{ch}$ and $R_{ch}$ from $H_{ch}$. Lastly, in Step\# 3, contrastive learning is used as discriminative task on the augmented views of $H_{ch}$ and $R_{ch}$ to learn the final representations. Here, the weights for $\phi$ and $\psi$ are initialised to those learnt during cross-stain prediction (i.e.\ Step \#2), thereby combining the strength of generative and discriminative learning.}
    \label{fig:csco}
\end{figure}

\begin{figure}[!tb]
\centering \footnotesize
\settoheight{\tempdima}{\includegraphics[width=0.15\textwidth]{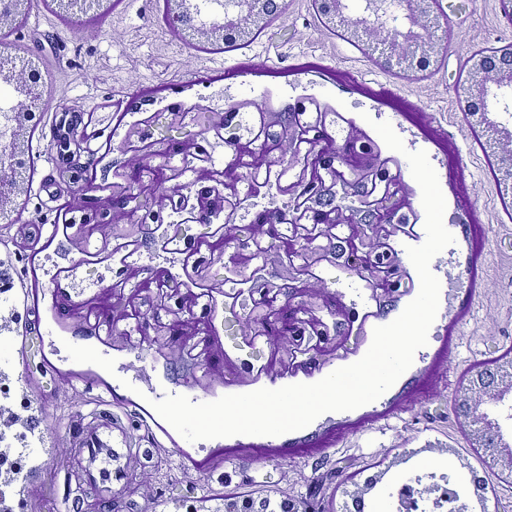}}%
\begin{tabular}{c@{ }c@{ }c@{ }c@{ }c@{ }c@{ }}
& PAS & Jones H\&E & CD68 & Sirius Red & CD34 
\\
\rowname{Stains} & 
\includegraphics[width=0.15\textwidth]{methods/stain_separation/stains/02.png} &
\includegraphics[width=0.15\textwidth]{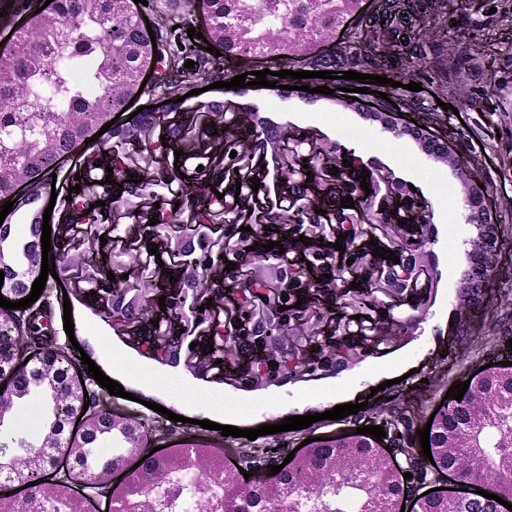} &
\includegraphics[width=0.15\textwidth]{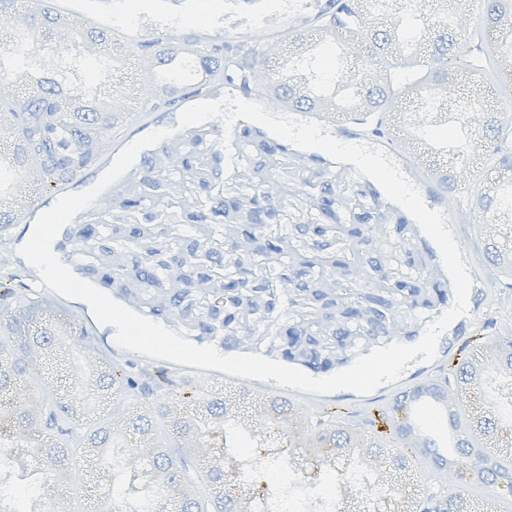} &
\includegraphics[width=0.15\textwidth]{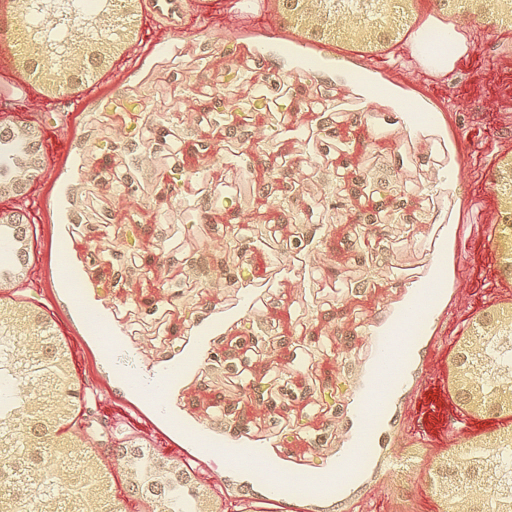} &
\includegraphics[width=0.15\textwidth]{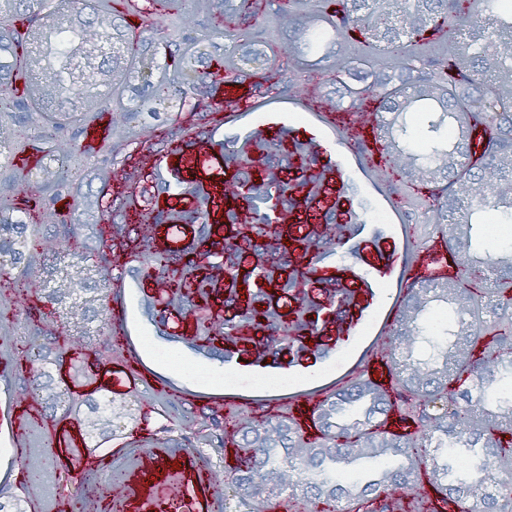}
\\
\rowname{$H_{ch}$} & \includegraphics[width=0.15\textwidth]{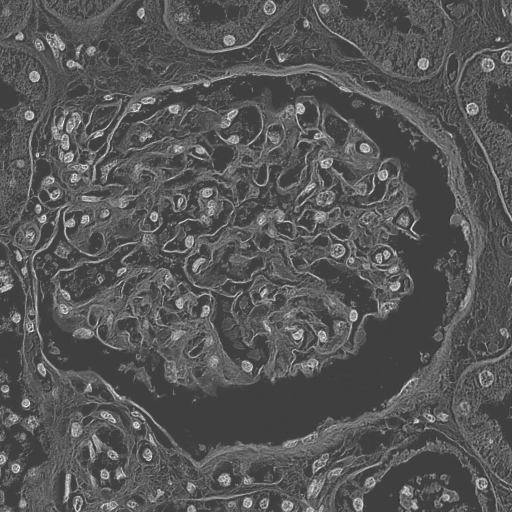} &
\includegraphics[width=0.15\textwidth]{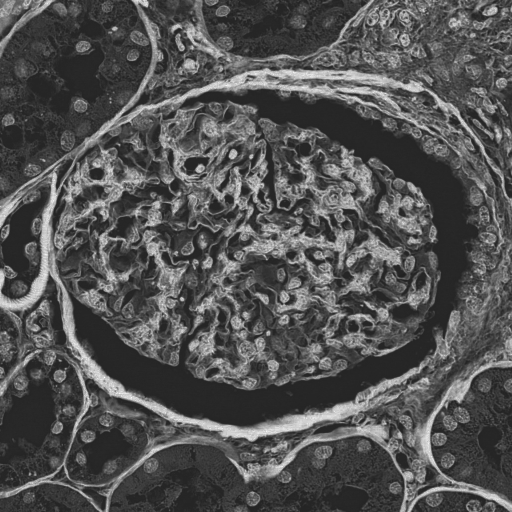} &
\includegraphics[width=0.15\textwidth]{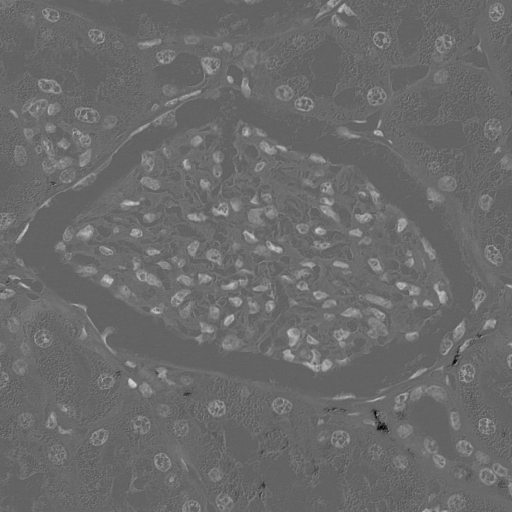} &
\includegraphics[width=0.15\textwidth]{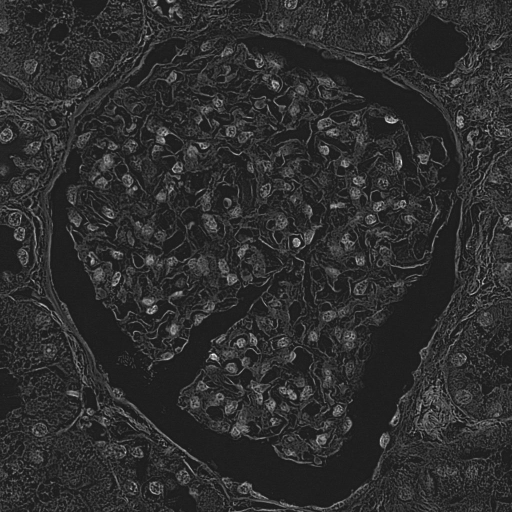} &
\includegraphics[width=0.15\textwidth]{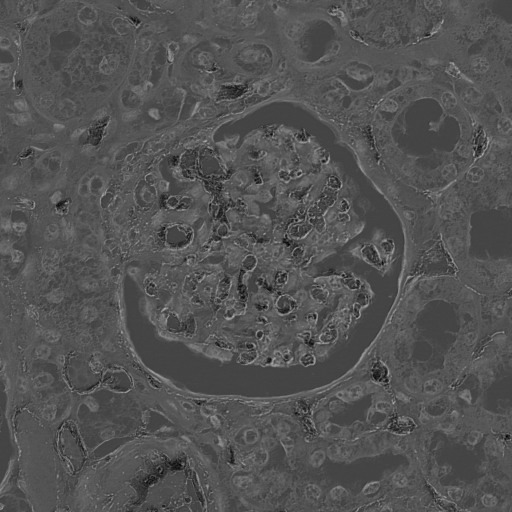}
\\
\rowname{$R_{ch}$} & 
\includegraphics[width=0.15\textwidth]{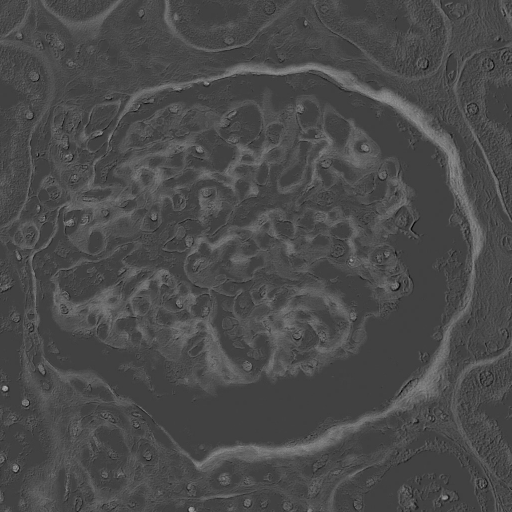} &
\includegraphics[width=0.15\textwidth]{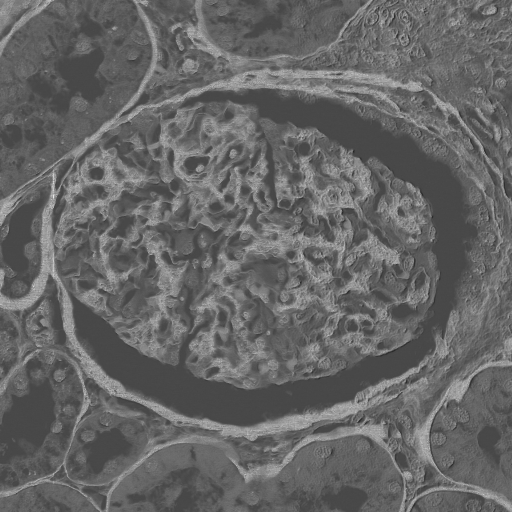} &
\includegraphics[width=0.15\textwidth]{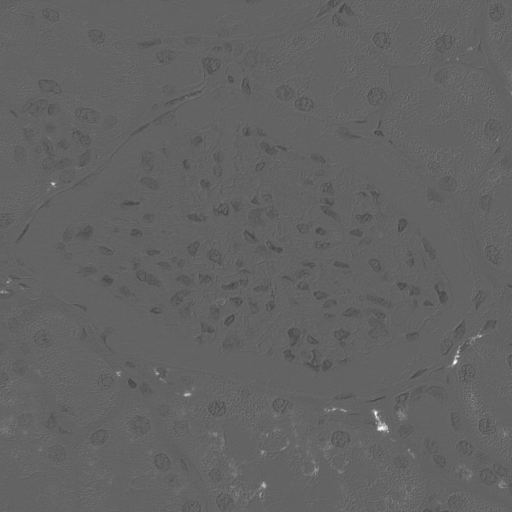} &
\includegraphics[width=0.15\textwidth]{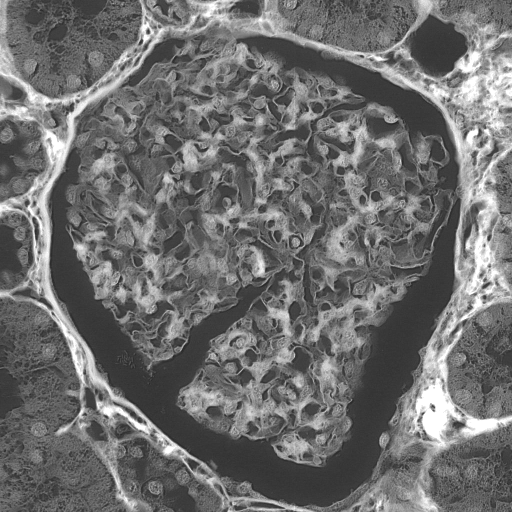} &
\includegraphics[width=0.15\textwidth]{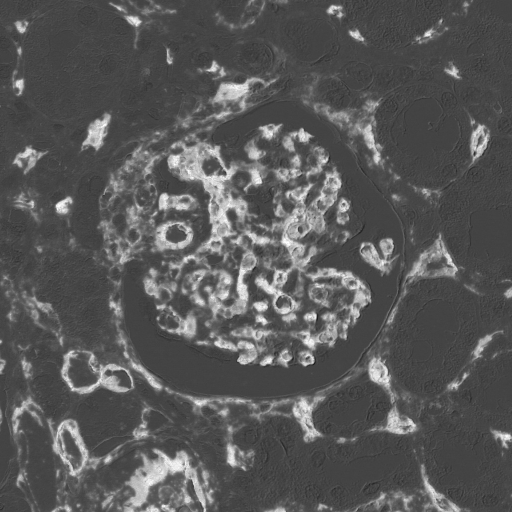} 
\end{tabular}
\caption{Visualisation of Haematoxylin ($H_{ch}$) and Residual ($H_{ch}$) channels extracted from each of the stains used in this study. }
\label{fig:stain_separation_better_visualisation}
\end{figure}

To address these limitations and extend the applicability of CS-CO across multiple stainings, we propose to modify its stain-separation strategy as outlined in Figure \ref{fig:csco}. Particularly, we exploit the fact that Haematoxylin is often used as a counterstain in histopathology, and therefore exists in many stains. This was first exploited by \citet{lampert2019strategies} as a strategy for stain invariant segmentation. Here, however, we use it to extract a common Haematoxylin channel, $H_{ch}$, which highlights cell nuclei, via image deconvolution \cite{ruifrok2001quantification}. The remaining information is retained as a `Residual' channel ($R_{ch}$) as illustrated in Fig.\ \ref{fig:csco}, step \#1, and Fig.\ \ref{fig:stain_separation_better_visualisation}, capturing tissue structure highlighted by the other stain components, such as glycogen, collagen, macrophages, and endothelial cells, etc, depending on the staining used. Therefore, all stain (containing $H_{ch}$) can be included by modelling them as $H_{ch}$ and $R_{ch}$. In the rest of the article, we refer to this modified version of CS-CO as HR-CS-CO.

Moving forward, in the cross-stain prediction stage, two separate auto-encoders H2R and R2H are trained as shown in Fig.\ \ref{fig:csco} (Step \#2). H2R learns to predict $R_{ch}$ from $H_{ch}$, and R2H performs the inverse task. Both share the same architecture but have different weights. For simplicity, $\phi_{h2r}$ and $\psi_{h2r}$ is used to represent the encoder and decoder for H2R (and similarly for R2H). Additionally, the combination of $\phi_{h2r}$ and $\phi_{r2h}$, and $\psi_{h2r}$ and $\psi_{r2h}$, are denoted as the HR encoder and decoder respectively. The mean square error (MSE) loss is computed to evaluate the dissimilarity between the real ($H_{ch}, R_{ch}$) and predicted ($H^{*}_{ch}, R^{*}_{ch}$) images, such that 
\begin{equation}
    \mathcal{L}_{cs}=\left(H_{ch} - H^{*}_{ch}\right)^{2} + \left(R_{ch} - R^{*}_{ch}\right)^{2},
\label{eq:csco_cs}
\end{equation}
where $R^{*}_{ch}=\psi_{h2r}(\phi_{h2r}(H_{ch}))$ and $H^{*}_{ch}=\psi_{r2h}(\phi_{r2h}(R_{ch}))$. This makes the two-branched auto-encoder sensitive to low-level features \cite{yang2022cs}.

Next, contrastive learning is used in Step\# 3 (as a final step) to learn high-level discriminative features. The model is reorganised into a Siamese architecture \cite{chen2021exploring}, consisting of the HR encoder ($\phi$), a projection head ($g$), and a prediction head ($q$), in which the parameters of the two branches are shared. Both $g$ and $q$ are multi-layer perceptrons (MLP) with the same architecture. To prevent mode collapse, the HR decoder ($\psi$), is retained in one branch as a non-trainable regulator. Instead of employing random initialisation, the weights for $\phi$ and $\psi$ are initialised to those learnt during cross-stain prediction (i.e.\ Step \#2), thereby combining the strength of general low-level and discriminative high-level features.  

\begin{figure}[!tb]
    \centering
    \includegraphics[width=0.6\textwidth]{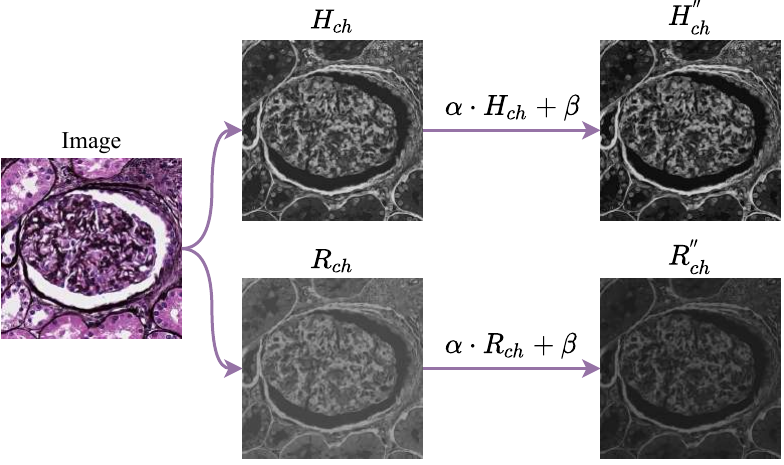}
    \caption{Stain-variation augmentation. From left to right: the process begins by decomposing an Image into its corresponding Haematoxylin ($H_{ch}$) and Residual ($R_{ch}$) channels. Subsequently, each channel undergoes individual modification using a random factor $\alpha$ and bias $\beta$. The modified version are represented as $H''_{ch}$, and $R''_{ch}$.}
     \label{fig:stain_variation}
\end{figure}
During contrastive learning, $H_{ch}$ and $R_{ch}$ are extracted from a given input image $x$ to give ($x_{h}, x_{r}$) and augmentation ($f_{aug}$) is applied to generate two distinct augmented views of each data sample: ($x'_{h}, x'_{r}$) and ($x''_{h}, x''_{r}$). $f_{aug}$ includes various augmentations such as flipping, random cropping and resizing, and Gaussian blur. Since the images are grey-scale, colour-based augmentations are not used, however stain variation is included as an additional augmentation, where the pixel intensities of each extracted channel are modified by a random factor $\alpha \in [-0.25, 0.25]$ and bias $\beta \in [-0.05, 0.05]$. These values were chosen as they result in realistic output, see Fig.\ \ref{fig:stain_variation}.

Subsequently, each augmented pair is fed into the Siamese network and encoded by $\phi_{h2r}$ and $\phi_{r2h}$. The outputs are pooled and concatenated to form a single vector, which is processed by $g$ to obtain ($z{'}, z{''}$) and $q$ to obtain ($w{'}, w{''}$). The symmetrised loss is
\begin{equation}
    \mathcal{L}_{co}=\frac{1}{2}\|\bar{w}' - \bar{z}''\|^2_2 + \frac{1}{2}\|\bar{w}'' - \bar{z}'\|^2_2,
\label{eq:csco_co}
\end{equation}
where $\bar{w}'$, $\bar{w}''$, $\bar{z}'$, and $\bar{z}'$ are the $\ell_{2}$ normalised versions of $w'$, $w''$, $z'$, and $z''$, respectively. This encourages $w'$ and $w''$ to be similar to $z''$ and $z'$ (respectively). Before computing the loss, the stop-gradient ($sg$) is applied to $z'$ and $z''$. This $sg$ step introduces a necessary asymmetry in gradient flow, which, when combined with the symmetrised loss, allows robust and diverse feature learning. During contrastive learning, the frozen pre-trained HR decoder ($\psi$) continues to use the outputs of the HR encoder ($\phi$) for image reconstruction. To avoid collapse, the HR encoder must maintain the necessary information for image reconstruction, by satisfying Equation \ref{eq:csco_cs}. As a result, the total loss is formulated as $\mathcal{L}_{csco}=\mathcal{L}_{cs} + \gamma\mathcal{L}_{co}$, where $\gamma$ is the weight coefficient.

\subsection{Downstream Segmentation Tasks} \label{sec:methods::subsec:downstream_tasks}
\begin{figure}[!tb]
    \centering
    \includegraphics[width=0.6\textwidth]{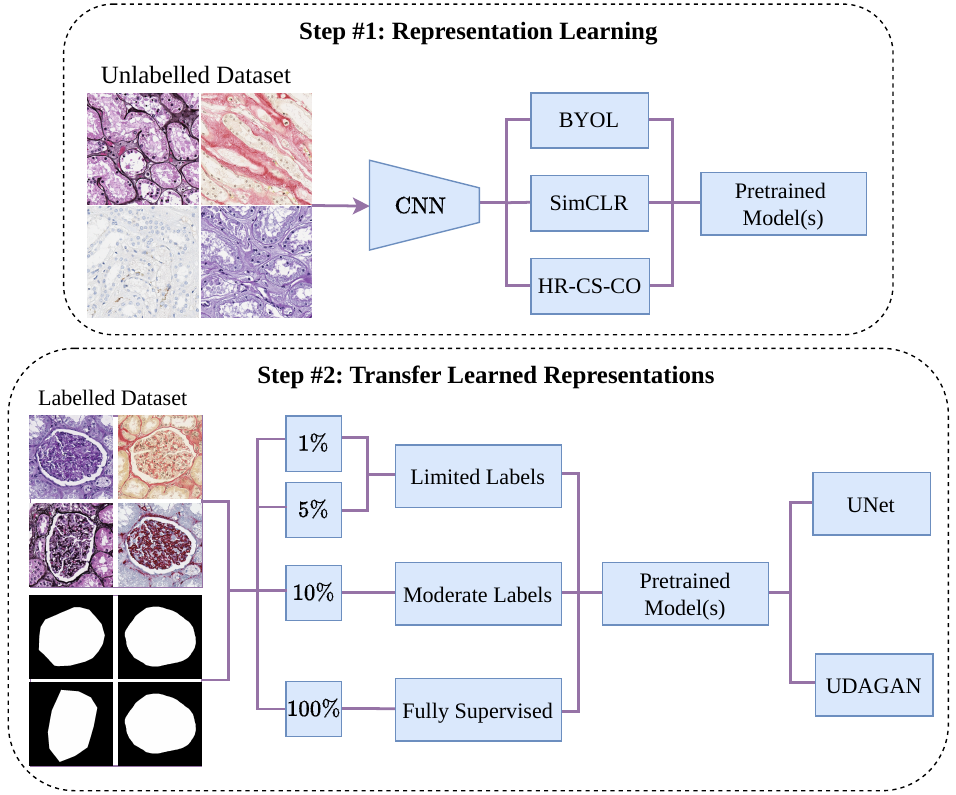}
    \caption{Workflow of our study. Step \#1: Different SSL methods are applied to learn representations from a large unlabelled dataset. Step \# 2: The learned representations are then refined by fine-tuning on a small labelled data for different histopathology related segmentation tasks.}
    \label{fig:paper_workflow}
\end{figure}

Once a model has been pre-trained using one of the self-supervised pre-training methods mentioned above, it can be fine-tuned for various downstream tasks. The primary aim of this study is to investigate the use of these pre-training methods to reduce the need of labels for several segmentation tasks, such as UNet \cite{ronneberger2015u}, and UDAGAN \cite{vasiljevic2021towards}, as shown in Fig.\ \ref{fig:paper_workflow}. The original architectures as proposed by the authors were used, with details provided below. Additional details on combining these with pre-training are presented in Section \ref{sec:training}.

\subsubsection{UNet} \label{methods:unet}
\begin{figure}[!tb]
    \centering
    \includegraphics[width=0.8\textwidth]{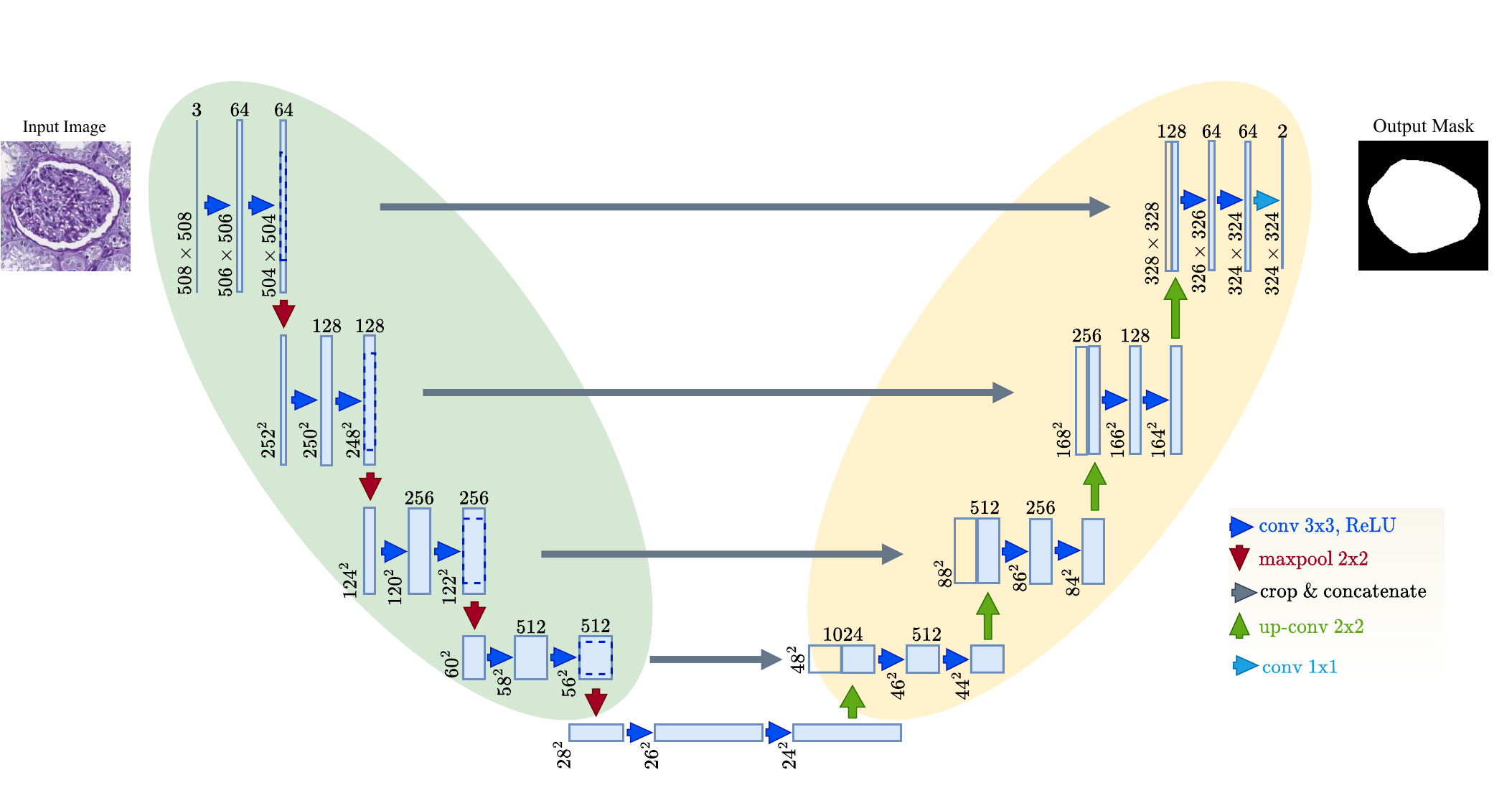}
    \caption{Overview of the UNet architecture.}
    \label{fig:unet}
\end{figure}
UNet \cite{ronneberger2015u} is a highly effective CNN architecture, see Fig.\ \ref{fig:unet}, that has shown remarkable efficacy in segmenting biomedical images, specifically for glomeruli segmentation \cite{de2018automatic, lampert2019strategies}. It adopts an encoder-decoder structure, forming a U-shaped network, which effectively handles both local and global information. The encoder path, also known as the contracting path, comprises repetitive blocks, each encompassing two consecutive $3\times3$ convolutions followed by ReLU activation and a max-pooling layer. Conversely, the decoder path, or expanding path, gradually upsamples the feature maps using $2\times2$ transposed convolution layers. Subsequently, the corresponding feature map from the contracting path is cropped and concatenated with the up-sampled feature map, followed by two consecutive $3\times3$ convolutions and a ReLU activation. Finally, a $1\times1$ convolution is applied to reduce the feature map to the desired number of channels (classes), generating the segmentation map. The cropping step is necessary since pixels at the edges contains less contextual information and therefore should be discarded.

\subsubsection{UDAGAN} \label{methods:mds1_udagan}
UDAGAN \cite{vasiljevic2021towards} is an advanced stain transfer based multi-stain segmentation approach, that use labels from the source ($S$) stain, which is PAS 
in our case, and eliminate the need for labels in the target ($T$) stains (herein these are Jones H\&E, Sirius Red, CD68, CD34). Stain transfer is achieved using a cycle-consistent generative adversarial network (CycleGAN) \cite{zhu2017unpaired}. Specifically, a separate CycleGAN model ($M^{S \leftrightarrow T}_\text{cycGAN}$) is trained to translate between the $S$ and each $T$ stain, i.e.\ from $S \rightarrow T$ and $T \rightarrow S$, see Fig.\ \ref{fig:cyclegan}. The architectural descriptions for CycleGAN are detailed in \cite{zhu2017unpaired}. 

\begin{figure}[!tb]
    \centering
    \includegraphics[width=\textwidth]{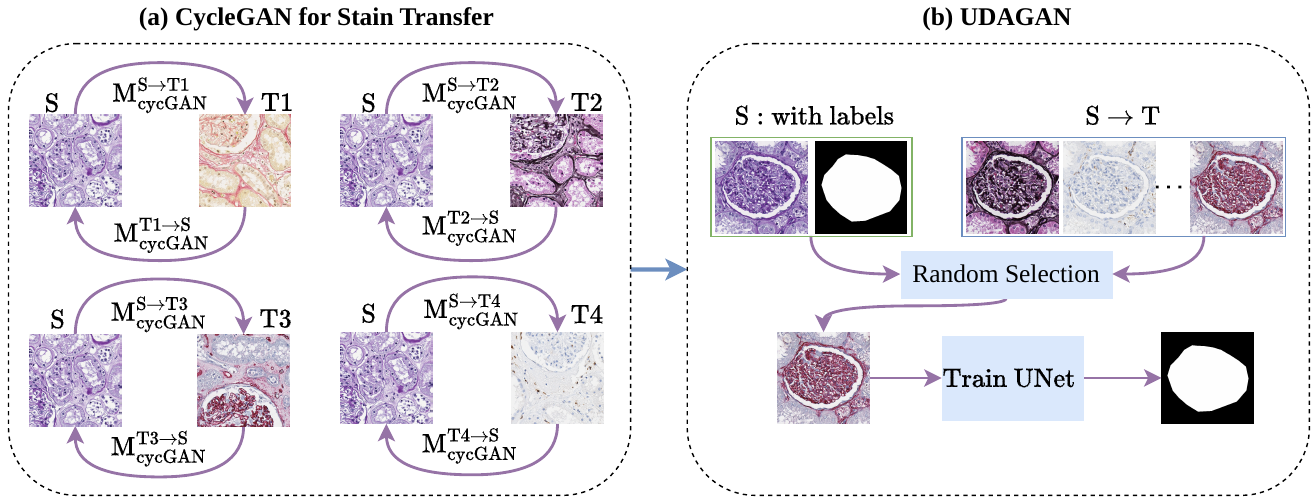}
    \caption{Overview of UDAGAN architecture.}
    \label{fig:udagan_full}
    \extralabel{fig:cyclegan}{(a)}
    \extralabel{fig:udagan}{(b)}
\end{figure}


Once the CycleGAN models ($M^{S \rightarrow T}_\text{cycGAN}$) are trained, they are used to augment the labelled source stain by randomly translating images to one of the target stains. Since translation does not change the image's structure, see Fig.\ \ref{fig:udagan}, it's ground truth is still valid. As such, samples from all available stains, but that are annotated in the source stain, are presented to the UNet, resulting in a single stain-invariant segmentation model---UDAGAN---that can be directly applied to multiple stains.

\section{Experimental Setup}\label{sec:experimental_setup}
\subsection{Dataset} \label{sec:data}
The findings presented herein were evaluated for kidney glomeruli segmentation using a private in-house histopathology dataset provided by the Hannover Medical School. Tissue samples were collected from $10$ patients who had tumor nephrectomy due to renal carcinoma. The renal tissue was selected as distant as possible from the tumors to represent largely normal renal glomeruli. However, certain samples exhibited varying degrees of pathological modifications, such as complete or partial displacement of functional tissue by fibrotic changes (``scerosis'') indicating normal age-related changes or the renal effects of general cardiovascular comorbidity (e.g.\ cardial arrhythmia, hypertension, arteriosclerosis). Using the Ventana Benchmark Ultra automated staining tool, the paraffin-embedded samples were sliced into 3\si{\micro\meter} thick sections and stained with either Jones H\&E, PAS, Sirius Red, and two immunohistochemistry markers (CD34, and CD68). An Aperio AT2 scanner was used to capture whole slide pictures at 40 magnification (a resolution of 0.253 \si{\micro\meter} / pixel), resulting in WSIs ranging from 37k$\times$25k to 107k$\times$86k pixels. Pathologists annotated and verified all of the glomeruli in each whole slide image (WSI) by labelling them with Cytomine \cite{maree2016collaborative}. The dataset was split into 4 training, 2 validation, and 4 test patients. The level-of-detail used is 1 (corresponding to 20$\times$ magnification) with a patch size of 508$\times$508 pixels, as to contain a glomeruli with its surrounding area.
\begin{table}[tb]
\caption{Training data with different percentages of labelled glomeruli for each staining.}
\label{tab:fraction_glomeruli}
\centering \scriptsize
\begin{NiceTabular}{cccccc}
\toprule
\Block{2-1}{\% of Labels} & \Block{1-5}{Stainings} \\
\cmidrule{2-6}
& PAS & Jones H\&E & CD68 & Sirius Red & CD34  \\
\midrule
1\% & 6 & 5 & 5 & 6 & 5 \\
\midrule
5\% & 33 & 31 & 26 & 32 & 28 \\
\midrule
10\% & 66 & 62 & 52 & 65 & 56 \\
\midrule
100\% & 662 & 621 & 526 & 651 & 565 \\
\bottomrule
\end{NiceTabular}
\end{table}

\subsubsection{Self-Supervised Pre-Training} \label{sec:data::subsec:ssl}
The dataset for self-supervised pre-training is extracted from the training and validation WSIs in an unsupervised manner (uniformly sampled). To be representative of each patient, 
$15,000$ (training) and $1,000$ (validation) patches were extracted
from each stain, resulting in $75,000$ training and $5,000$ validation patches. 

\subsubsection{Downstream Segmentation Tasks} \label{sec:data::subsec:segmentation}
Glomeruli segmentation is framed as a two class problem: glomeruli (pixels belonging to glomeruli), and tissue (pixels outside a glomerulus). The performance of the trained models is evaluated by segmenting the full test patient  WSIs from each stain. The number of glomeruli present in the test stainings are: PAS - 1092; Jones H\&E - 1043; Sirius Red - 1049; CD34 - 1019; CD68 - 1046.
\begin{description}[leftmargin=0pt]
\item[UNet --]
Following \citet{ciga2022self}, multiple splits of the overall dataset were created that contains different percentages of labelled data (1\%, 5\%, 10\%, 100\%) taken from the training patients of each stain (see Table \ref{tab:fraction_glomeruli}). Additionally, seven times more tissue (i.e.\ non-glomeruli) patches were included to account for the variability observed in non-glomeruli tissue. Similar to \citet{stacke2021learning}, a validation dataset is used to select the best model based on the loss. The number of glomeruli present in the validation stainings are: PAS - 588; Jones H\&E - 590; Sirius Red - 576; CD34 - 595; CD68 - 521.  

\item[UDAGAN --]
To train the respective CycleGAN models, $5000$ patches are randomly extracted from the training WSIs of each stain (PAS being the source, and the rest the targets). 
Additionally, when fine-tuning the pre-trained UDAGAN models, 
the same splits as used for the UNet are preserved, but only the samples from the PAS stain are used, see Table \ref{tab:fraction_glomeruli} ($1^{\text{st}}$ column).
\end{description}

\subsection{Training Details} \label{sec:training}
\subsubsection{Self-Supervised Pre-training} \label{sec:training::subsec:ssl}
SimCLR and BYOL use CNN based encoders however, CS-CO relies on a CNN-based auto-encoder. The UNet offers SOTA performance in glomeruli segmentation \cite{de2018automatic, lampert2019strategies} and therefore its encoder was used for SimCLR and BYOL and encoder and decoder for CS-CO. In each of these networks, the extracted representations are subsequently projected into a lower-dimensional space using a multi-layer perceptron (MLP). The best trained model is selected 
based on the self-supervised validation dataset. Each SSL network was trained once due to computational and time constraints. The training details for each self-supervised pre-training method are detailed below.

\begin{description}[leftmargin=0pt]
\item[SimCLR:]
The training setup proposed in the original paper \cite{chen2020simple} was used. Recently, \citet{stacke2021learning} has shown that smaller batch sizes are preferable when using SimCLR in histopathology, particularly when dealing with few classes, and therefore a batch size of 256 was used, since this reduces the risk of false negatives. This also allowed the higher resolution (i.e.\ $508\times508$ pixels) of histopathological images to be used. Following \cite{stacke2021learning}, we trained  SimCLR  for $200$ epochs.

\item[BYOL:] A similar training procedure as described in the original BYOL paper \cite{grill2020bootstrap} was used. The absence of negative samples in BYOL's training paradigm allows it to sustain performance parity with SimCLR despite using smaller batch sizes. A batch size of $256$ was chosen and the model was trained for $200$ epochs. Since BYOL can be susceptible to poor initialisation, 
the encoder uses batch-normalisation (BN). 

\item[HR-CS-CO:]
Since the concentration of the $H_{ch}$ can vary between different stainings, we train separate HR-CS-CO models for each. The two training stages (see Section \ref{sec:methods::subsec:ssl}) have the following training strategies: Cross-stain prediction, the model is trained for $100$ epochs with a batch size of $32$ using the Adam optimiser with initial learning rate (LR) of $0.001$, which, based on the validation loss and a patience of $10$ epochs, is reduced by a factor of $0.1$; Contrastive learning, the model is trained again for $50$ epochs and a batch size of $128$ using the Adam optimiser with LR of $0.001$ and a weight decay of $10^{-6}$. To prevent over-fitting, early stopping is used in both stages.
\end{description}

\subsubsection{Downstream Segmentation Tasks} \label{sec:training::subsec:segmentation}
As is common, the validation loss (see Section \ref{sec:data::subsec:segmentation}) is used to select the final model 
\cite{stacke2021learning, yang2022cs, ciga2022self}. This is, however, a hindrance in histopathology (and medical imaging in general) since it requires additional labels. 
Thus, additional experiments were performed without a validation set by selecting the best model based on the training loss.
The training details for both downstream tasks (UNet, and UDAGAN) and the strategy to combine the pre-trained features from each of the self-supervised pre-trained method are detailed below.

\begin{description}[leftmargin=0pt]
\item[UNet:]
The UNet is trained for 250 epochs (following \citet{vasiljevic2021towards}) using a batch size of $8$ and a learning rate of $0.0001$. All patches are standardised to $[0, 1]$ and normalised by the mean and standard deviation of the training set. The same augmentation as used by \citet{lampert2019strategies} are applied with an independent probability of $0.5$ (batches are augmented `on the fly'), that is: 
elastic deformation  ($\sigma = 10$, $\alpha = 100$); random rotation in the range $\intervaldeg{0}{180}$, random shift sampled from $\interval{-205}{205}$ pixels, random magnification sampled from $\interval{0.8}{1.2}$, and horizontal/vertical flip; additive Gaussian noise with $\sigma \in \interval{0}{2.55}$; Gaussian filtering with $\sigma \in \interval{0}{1}$; brightness, colour, and contrast enhancements with factors sampled from $\interval{0.9}{1.1}$; stain variation by colour deconvolution \cite{tellez2018whole}, $\alpha$ sampled from $\interval{-0.25}{0.25}$ and $\beta$ from $\interval{-0.05}{0.05}$.

To integrate the benefits of the knowledge gained from self-supervised pre-training, the encoder component of the UNet architecture is initialised with the weights learned during pre-training (i.e.\ with SimCLR, BYOL, or CS-CO). Five different repetitions of the UNet model were trained for each stain and for each split of labelled data. 


\item[UDAGAN:]
Training the UDAGAN involves both UNet and CycleGAN networks. Specifically, in the first step, a CycleGAN network is trained for each target stain, enabling the translation from $S \rightarrow T$. To train these CycleGAN models, the network architecture and loss weights ($w_{cyc}=10, w_{id}=5$) are taken from the original paper \cite{zhu2017unpaired}. To deal with large patch sizes (i.e.\ above $256\times256$ pixels), a translation network with nine ResNet blocks is employed. The model is trained for $50$ epochs, with a LR of $0.0002$ using the Adam optimiser, and a batch size of $1$. Starting from the $25^{\text{th}}$ epoch, the LR is linearly decayed to $0$, and the weights for cycle-consistency ($w_{cyc}$) and identity ($w_{id}$) are halved. In all experiments, the translation model from the final epoch is used. To reduce model oscillation, \citet{shrivastava2017learning}'s strategy is adopted that updates the discriminator using the $50$ previously generated samples. After training CycleGAN, a training patch from $S$ stain is translated to a randomly selected $T$ stain (with a probability of $\frac{N-1}{N}$, where $N$ is the number of stains) using the pre-trained CycleGAN network(s). Thus, all available stains (including the $S$ stain) are presented to the UNet model with equal probability, $\frac{1}{N}$, forcing the network to learn stain invariant features.
One UDAGAN model is trained for each split of the $S$ stain labels and repeated five times.
\end{description}

\section{Results} \label{sec:results}
In this section, the pre-trained models are evaluated for each downstream task (UNet and UDAGAN) in two different settings: fixed-features and fine-tuning. In the fixed-feature setting, the pre-trained weights are frozen to assess the quality of the learned representations from self-supervised pre-trained models, using the same hyperparameters as used for baseline models. When fine-tuning, the pre-trained weights are updated. To determine the optimal hyperparameters for fine-tuning, a separate hyperparameter study was conducted using 1\%, and 5\% splits of labelled data and the performance was evaluated on the validation set for each task and pre-training method. Five learning rate values, logarithmically spaced between 0.0001 and 0.1, were tested. Additionally, two different settings for weight decay were examined: one with a value of 10$^{-4}$ and one without any weight decay. The learning rate was reduced by a factor of 0.1 at the \nth{90} percentile of training. Based on these experiments, the best hyperparameters were selected, and the fine-tuned models were re-trained for all label splits. The F$_1$ score is used as the evaluation metric and the results are presented on a separate unseen test set (as outlined in Section \ref{sec:data::subsec:segmentation}).

Fully supervised models were trained to establish baselines for different label splits, including 100\% labels.  It was found that the fine-tuned models consistently outperform fixed-feature models, and therefore only fine-tuned results are shown here (fixed-feature results are in \ref{appen_sec:fixed_features_results}).

\begin{table}[!tb]
\centering \scriptsize
\addtolength{\leftskip} {-1.5cm}
\caption{A comparison of various self-supervised pre-training methods and respective baselines (randomly initialised without any pre-training) for the downstream tasks of UNet and UDAGAN using various splits of labelled data. For UNet, the labels have been used for all stains, while for UDAGAN, the labels for only source (PAS) stain are used. The evaluation is conducted on an independent, unseen test dataset using F$_{1}$ score. Each F$_{1}$ score is the average of five different training repetitions (standard deviations are in parentheses). The highest F$_{1}$ score for each stain, across different label splits, is in italics, while the overall highest F$_{1}$ score averaged across all stains is in bold.}
\label{table:finetune_results}
\begin{NiceTabular}{c:c:c:ccccc:c}
\toprule
\Block{2-1}{Downstream\\Task} & \Block{2-1}{Label\\Split} & \Block{2-1}{Method} & \Block{1-5}{Test Stain} & & & & & \Block{2-1}{Average}\\
\cmidrule{4-8}
& & & PAS & Jones H\&E & CD68 & Sirius Red & CD34 & \\
\midrule
\Block{16-1}{UNet} & \Block{4-1}{1\%} & None (Baseline) & 0.015 (0.031) & 0.000 (0.000) & 0.000 (0.000) & 0.000 (0.000) & 0.253 (0.059) & 0.054 (0.018)\\
& & SimCLR & \textit{0.673 (0.021)} & 0.519 (0.040) & 0.407 (0.015) & 0.472 (0.037) & 0.652 (0.018) & 0.544 (0.026)\\
& & BYOL & 0.660 (0.018) & \textit{0.635 (0.055)} & \textit{0.625 (0.042)} & \textit{0.561 (0.044)} & \textit{0.686 (0.030)} & \textbf{0.633 (0.038)}\\
& & HR-CS-CO & 0.154 (0.044) & 0.188 (0.067) & 0.048 (0.083) & 0.337 (0.082) & 0.463 (0.017) & 0.238 (0.058)\\
\cmidrule{2-9}
& \Block{4-1}{5\%} & None (Baseline) & 0.546 (0.084) & 0.593 (0.080) & 0.370 (0.188) & 0.707 (0.055) & 0.782 (0.041) & 0.600 (0.090)\\
& & SimCLR & \textit{0.852 (0.019)} & \textit{0.760 (0.017)} & 0.599 (0.039) & 0.618 (0.042) & \textit{0.802 (0.011)} & 0.726 (0.026)\\
& & BYOL & 0.768 (0.036) & 0.746 (0.076) & \textit{0.736 (0.033)} & \textit{0.721 (0.051)} & \textit{0.800 (0.047)} & \textbf{0.754 (0.049)}\\
& & HR-CS-CO & 0.756 (0.079) & 0.628 (0.086) & 0.533 (0.067) & 0.406 (0.067) & 0.707 (0.037) & 0.606 (0.067)\\
\cmidrule{2-9}
& \Block{4-1}{10\%} & None (Baseline) & 0.730 (0.017) & 0.792 (0.024) & 0.643 (0.053) & \textit{0.788 (0.022)} & 0.827 (0.063) & 0.756 (0.036)\\ 
& & SimCLR & \textit{0.867 (0.019)} & 0.813 (0.012) & 0.690 (0.057) & 0.696 (0.060) & \textit{0.838 (0.007)} & \textbf{0.781 (0.031)}\\ 
& & BYOL & 0.794 (0.047) & \textit{0.823 (0.054)} & \textit{0.729 (0.052)} & 0.722 (0.044) & 0.776 (0.057) & 0.769 (0.051)\\ 
& & HR-CS-CO & 0.807 (0.058) & 0.748 (0.098) & \textit{0.729 (0.040)} & 0.711 (0.074) & 0.791 (0.026) & 0.757 (0.059)\\
\cmidrule{2-9}
& \Block{4-1}{100\%} & None (Baseline) & \textit{0.894 (0.021)} & 0.840 (0.029) & 0.836 (0.031) & 0.865 (0.019) & \textit{0.888 (0.015)} & 0.865 (0.024)\\
& & SimCLR & 0.884 (0.003) & \textit{0.873 (0.007)} & 0.840 (0.011) & \textit{0.881 (0.007)} & 0.867 (0.027) & \textbf{0.869 (0.011)}\\
& & BYOL & 0.867 (0.009) & 0.842 (0.035) & 0.818 (0.036) & 0.847 (0.012) & 0.874 (0.021) & 0.850 (0.022)\\
& & HR-CS-CO & 0.843 (0.033) & 0.855 (0.015) & \textit{0.872 (0.006)} & 0.842 (0.023) & 0.870 (0.011) & 0.856 (0.018)\\

\midrule
\Block{12-1}{UDAGAN} & \Block{3-1}{1\%} & None (Baseline) & 0.000 (0.000) & 0.000 (0.000) & 0.000 (0.000) & 0.000 (0.000) & 0.000 (0.000) & 0.000 (0.000)\\
& & SimCLR & 0.477 (0.015) & 0.403 (0.025) & 0.261 (0.053) & 0.408 (0.010) & 0.518 (0.016) & 0.413 (0.024)\\
& & BYOL & \textit{0.647 (0.062)} & \textit{0.504 (0.083)} & \textit{0.401 (0.099)} & \textit{0.513 (0.088)} & \textit{0.598 (0.064)} & \textbf{0.533 (0.079)}\\ 
\cmidrule{2-9}
& \Block{3-1}{5\%} & None (Baseline) & 0.669 (0.038) & 0.498 (0.056) & 0.352 (0.056) & 0.618 (0.072) & 0.692 (0.024) & 0.566 (0.049)\\
& & SimCLR & 0.719 (0.018) & 0.616 (0.020) & 0.524 (0.014) & 0.632 (0.015) & 0.716 (0.015) & 0.641 (0.016)\\
& & BYOL & \textit{0.815 (0.027)} & \textit{0.730 (0.071)} & \textit{0.603 (0.028)} & \textit{0.732 (0.028)} & \textit{0.726 (0.055)} & \textbf{0.721 (0.042)}\\
\cmidrule{2-9}
& \Block{3-1}{10\%} & None (Baseline) & 0.816 (0.031) & 0.687 (0.014) & 0.614 (0.019) & 0.750 (0.069) & 0.770 (0.022) & 0.727 (0.031)\\ 
& & SimCLR & 0.781 (0.013) & 0.712 (0.013) & 0.606 (0.015) & 0.706 (0.026) & 0.768 (0.012) & 0.715 (0.016) \\  
& & BYOL & \textit{0.834 (0.035)} & \textit{0.767 (0.051)} & \textit{0.654 (0.040)} & \textit{0.742 (0.090)} & \textit{0.781 (0.037)} & \textbf{0.755 (0.051)}\\  
\cmidrule{2-9}
& \Block{3-1}{100\%} & None (Baseline) & \textit{0.901 (0.011)} & 0.856 (0.036) & 0.705 (0.031) & 0.873 (0.025) & 0.799 (0.035) & 0.827 (0.027)\\
& & SimCLR & 0.892 (0.008) & \textit{0.866 (0.018)} & \textit{0.777 (0.013)} & \textit{0.888 (0.015)} & \textit{0.844 (0.003)} & \textbf{0.853 (0.011)}\\
& & BYOL & 0.883 (0.019) & 0.854 (0.039) & 0.722 (0.051) & 0.818 (0.068) & 0.792 (0.036) & 0.814 (0.042)\\
\bottomrule
\end{NiceTabular}
\end{table}

The results presented in Table \ref{table:finetune_results} indicate that, in the majority of limited label scenarios (1\%, 5\%), the fine-tuned models consistently outperformed the baselines, while with moderate (10\%) and fully labelled (100\%) data, they result in similar or better performance across all stains.

On average, in the limited label cases, which are equivalent to 5--6 (1\%) and 26--33 (5\%) labelled glomeruli per stain, the fine-tuned UNet models significantly outperform the respective baseline UNet models (see last column). This outperformance is not uniform over all stains, however, notably Sirius Red and CD34 with 5\% labels do benefit from pre-training but not as considerably as the other stains. For some stains, it can be observed that pre-training with 100\% labels can even outperform the baseline fully supervised models, however, the benefits are not evident when averaging over all stains. As our goal is to find a labelling level that minimises labelling effort while maximising performance, 5\% labels offers a good balance between the two (10\% giving only a small increase in performance, while 1\% a considerable drop). At this level of labelling, a 11\% drop in performance is observed with BYOL pre-trained UNet in comparison to the fully (100\%) supervised model. This highlights that the number of labels required for training can be reduced by 95\%. If SSL had not been used in this case, a 26.9\% drop in performance would have been observed ($5^\text{th}$ row, last column of Table \ref{table:finetune_results}).  


This trend continues in the stain invariant UDAGAN model's results, where on average pre-training and fine-tuning with 1\% and 5\% labels (but in this case only from the source, PAS, stain) considerably outperforms the baselines in all stains. HR-CS-CO pre-training is not evaluated as UDAGAN is a stain-invariant single-model multi-stain segmentation approach and HR-CS-CO is trained separately for each stain. In this case, we observe a 10.6\% performance drop when fine-tuning with 5\% labels (and pre-training with BYOL) compared to the 100\% supervised baselines. If the model had been trained in a fully supervised manner with this amount of labels, a $26.1\%$ drop would have been observed, thus fine-tuning is able to minimise the impact of the lack of labels. 

\begin{figure}[!tb]
	\centering \scriptsize
	\settoheight{\tempdima}{\includegraphics[width=0.15\textwidth]{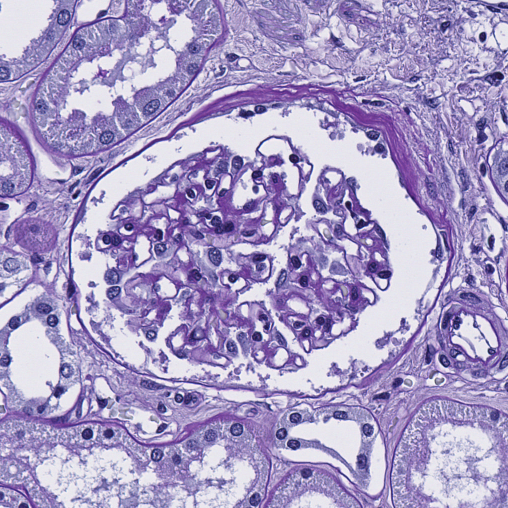}}%
	\begin{tabular}{c@{ }c@{ }c@{ }c@{ }c@{ }c@{ }c@{ }}
    & & PAS & Jones H\&E & CD68 & Sirius Red & CD34 
	\\
     & \rowname{Images} & 
     \includegraphics[width=0.15\textwidth]{results/images/IFTA_Nx_0018_02_glomeruli_patch_166.png} &
	\includegraphics[width=0.15\textwidth]{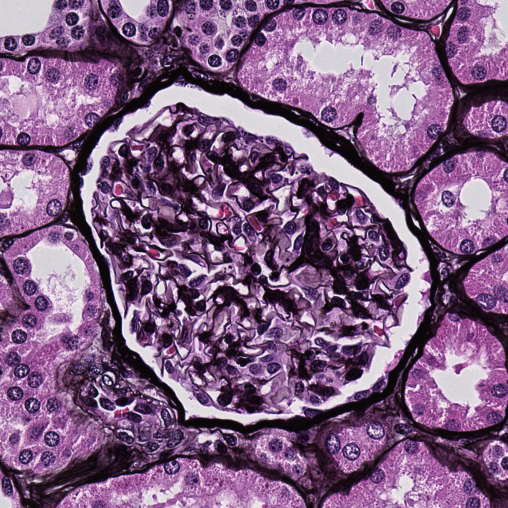} &
	\includegraphics[width=0.15\textwidth]{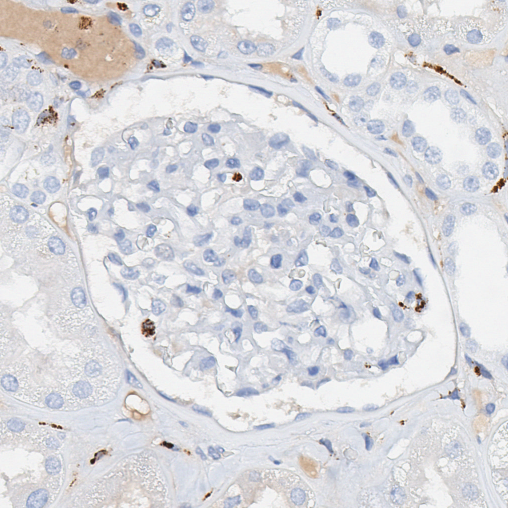} &
	\includegraphics[width=0.15\textwidth]{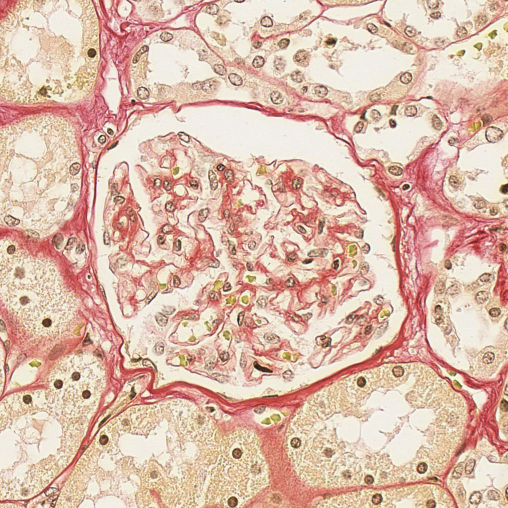} &
	\includegraphics[width=0.15\textwidth]{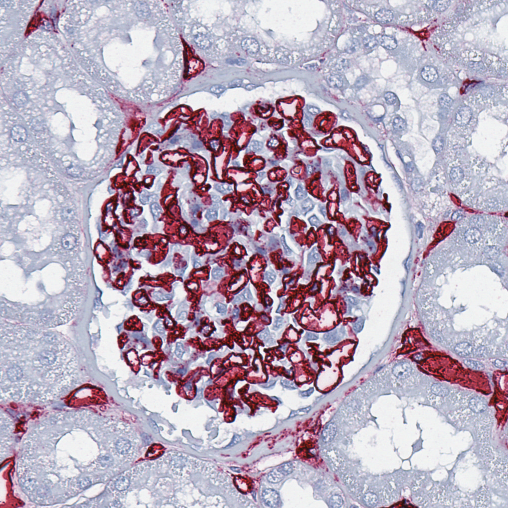} 
	\\
    & \rowname{GroundTruths} & \includegraphics[width=0.15\textwidth]{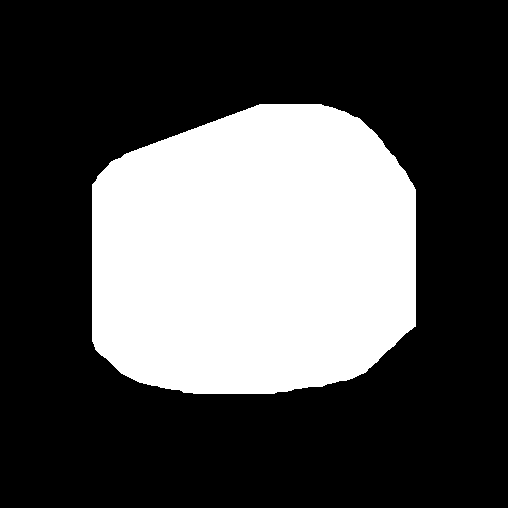} &
	\includegraphics[width=0.15\textwidth]{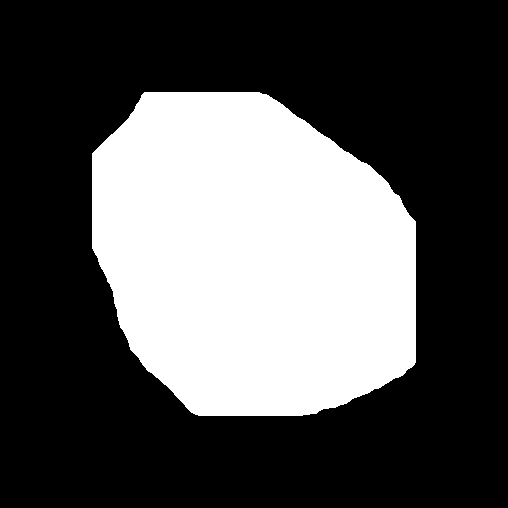} &
	\includegraphics[width=0.15\textwidth]{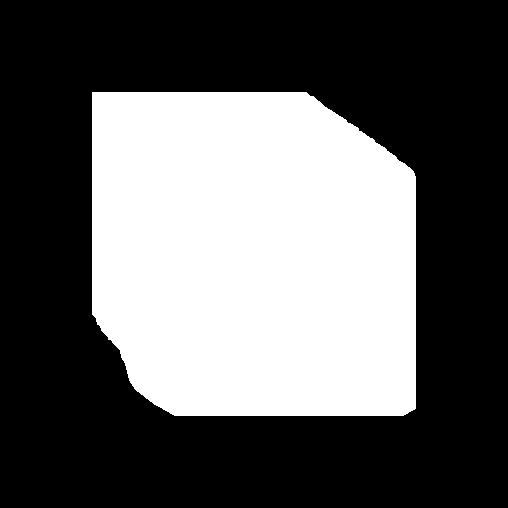} &
	\includegraphics[width=0.15\textwidth]{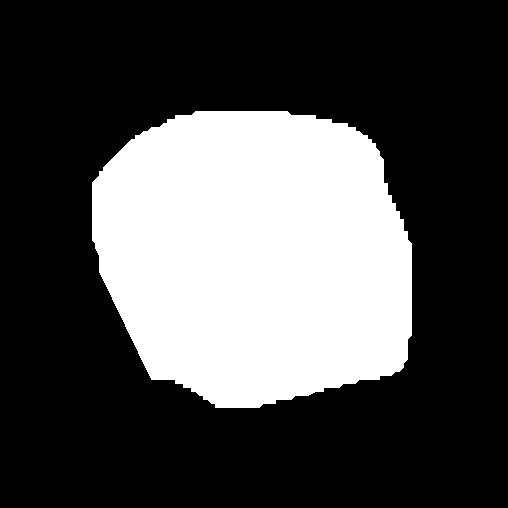} &
	\includegraphics[width=0.15\textwidth]{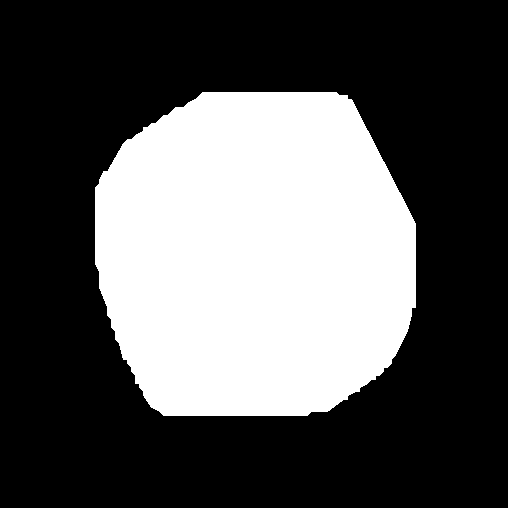}  
	\\
	\cline{1-7} \vspace{-1.5ex} 
	\\
    \multirow{3}{*}{\rotatebox[origin=c]{90}{UNet}} & \rowname{SimCLR} & 
	\includegraphics[width=0.15\textwidth]{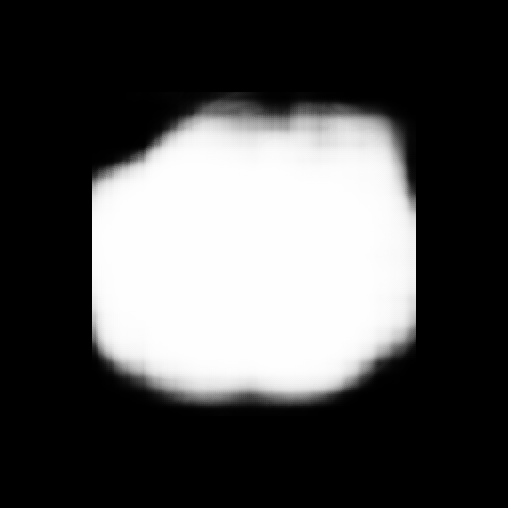} &
	\includegraphics[width=0.15\textwidth]{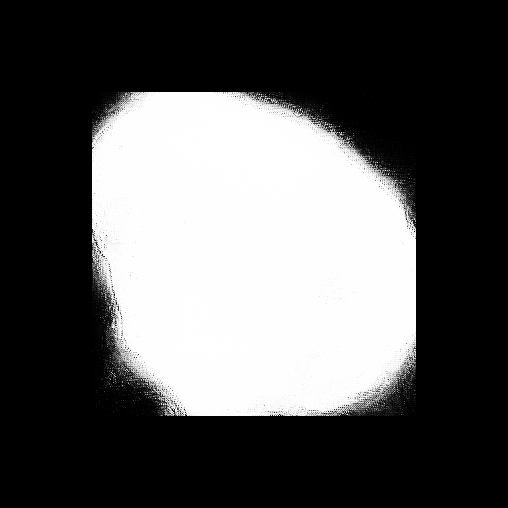} &
	\includegraphics[width=0.15\textwidth]{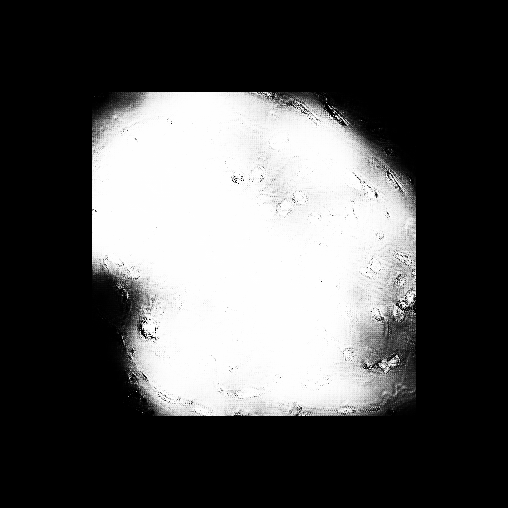} &
	\includegraphics[width=0.15\textwidth]{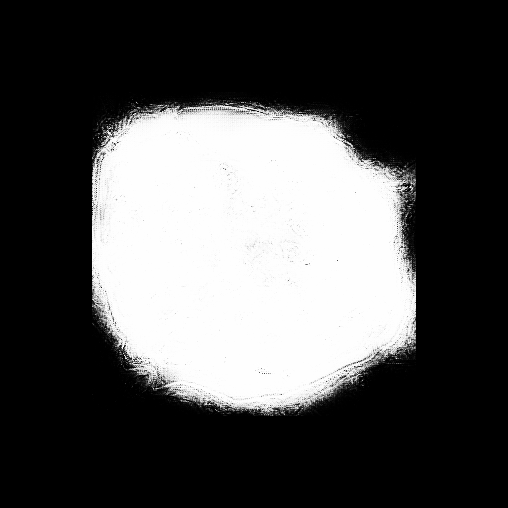} &
	\includegraphics[width=0.15\textwidth]{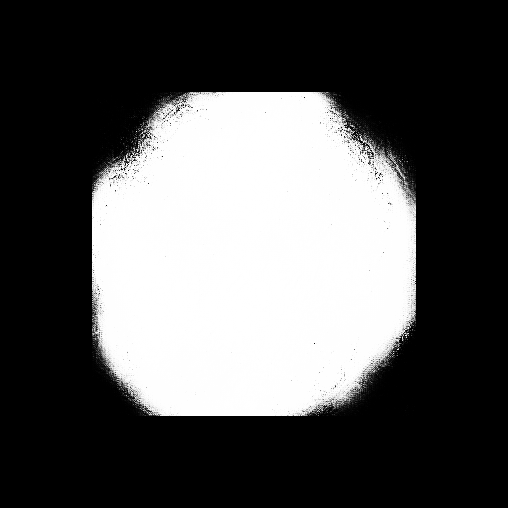} 
    \\
    & \rowname{BYOL} & 
	\includegraphics[width=0.15\textwidth]{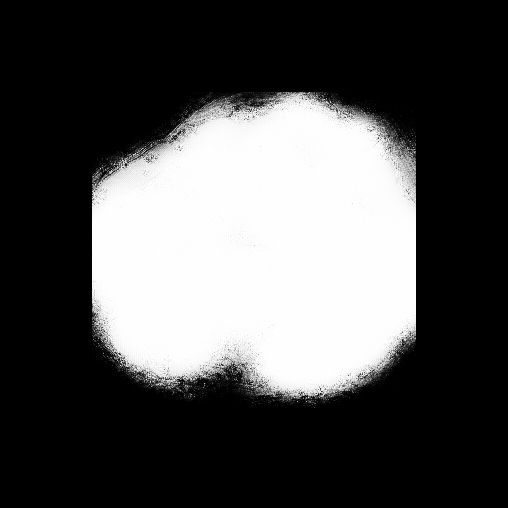} &
	\includegraphics[width=0.15\textwidth]{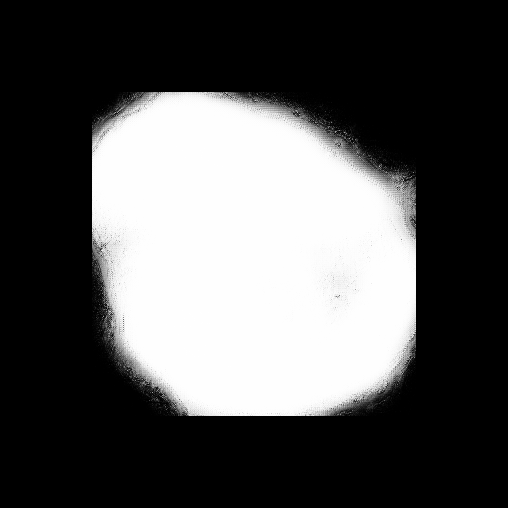} &
	\includegraphics[width=0.15\textwidth]{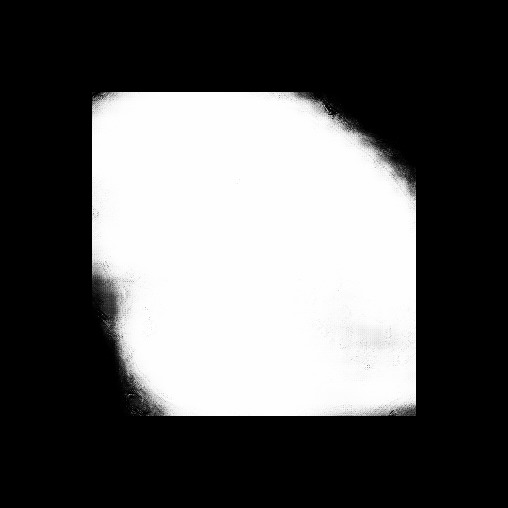} &
	\includegraphics[width=0.15\textwidth]{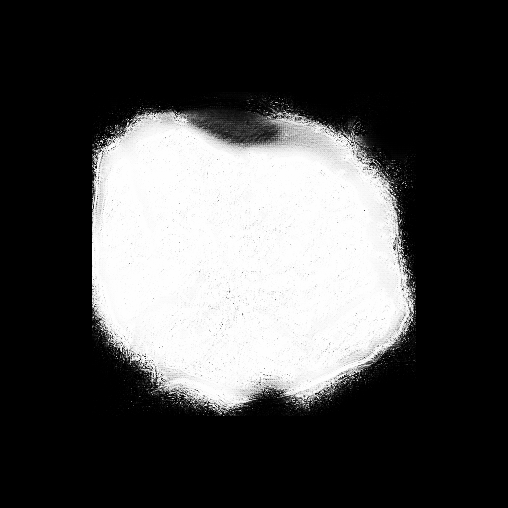} &
	\includegraphics[width=0.15\textwidth]{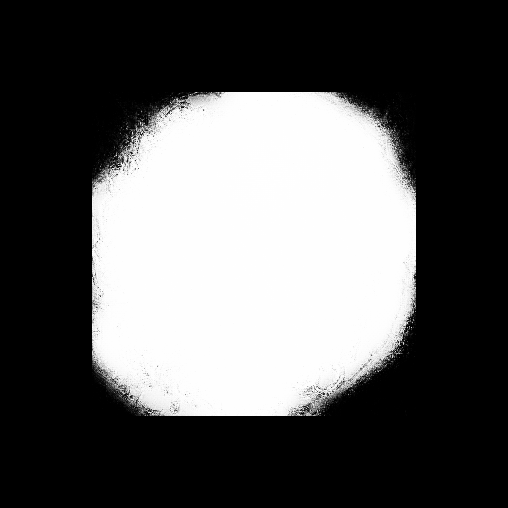} 
    \\
    & \rowname{HR-CS-CO} & 
	\includegraphics[width=0.15\textwidth]{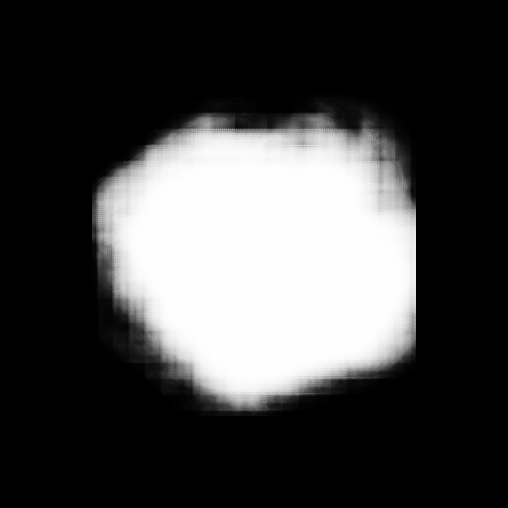} &
	\includegraphics[width=0.15\textwidth]{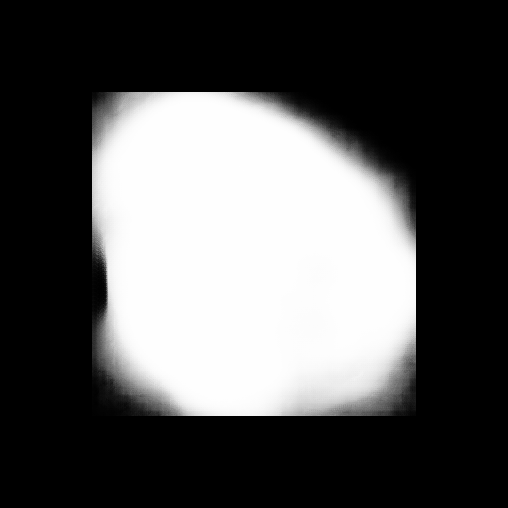}&
	\includegraphics[width=0.15\textwidth]{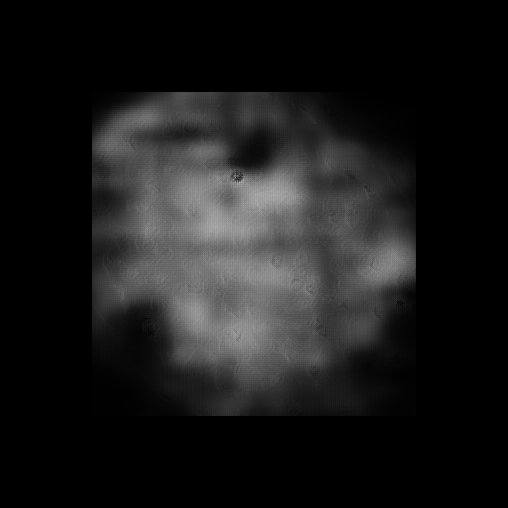}&
	\includegraphics[width=0.15\textwidth]{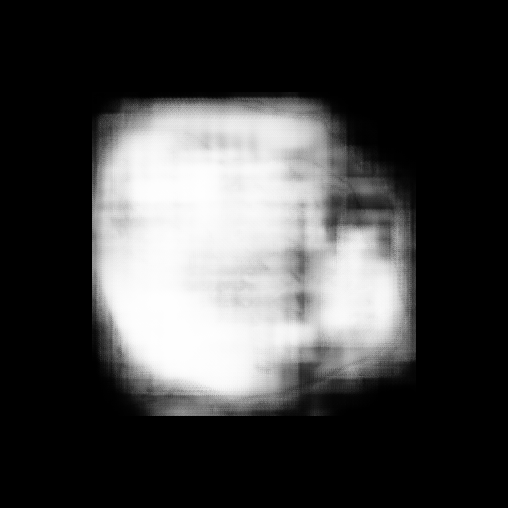} &
	\includegraphics[width=0.15\textwidth]{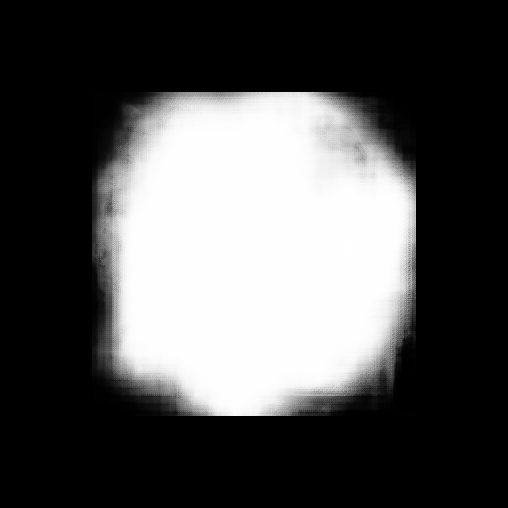}

    \\
    \cline{1-7} \vspace{-1.5ex} 
    \\
    \multirow{2}{*}{\rotatebox[origin=c]{90}{UDAGAN}} & \rowname{SimCLR} & 
	\includegraphics[width=0.15\textwidth]{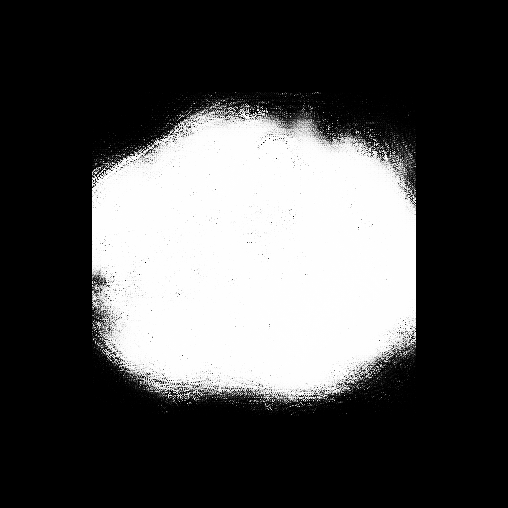} &
	\includegraphics[width=0.15\textwidth]{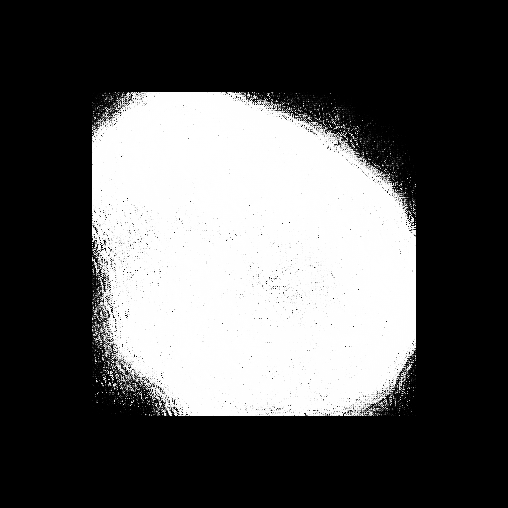} &
	\includegraphics[width=0.15\textwidth]{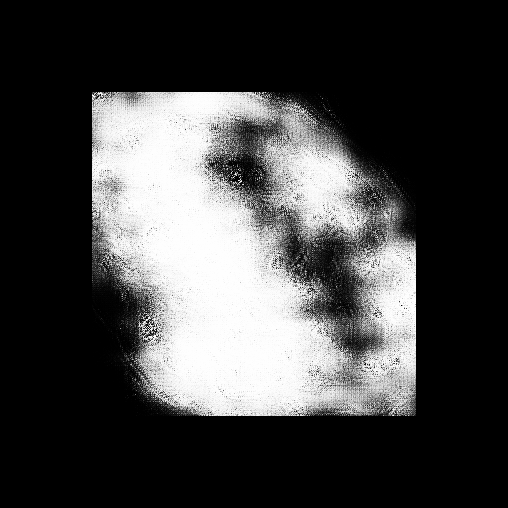} &
	\includegraphics[width=0.15\textwidth]{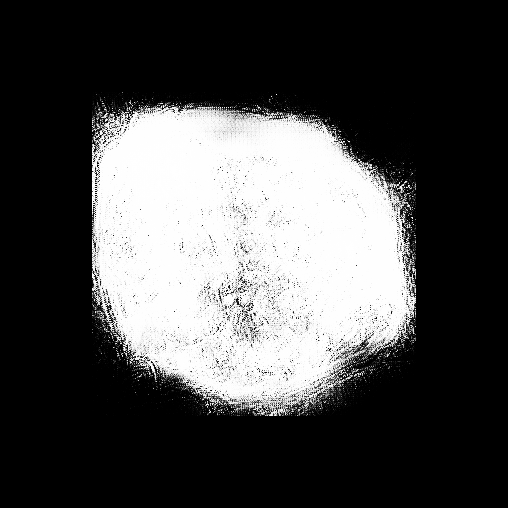} &
	\includegraphics[width=0.15\textwidth]{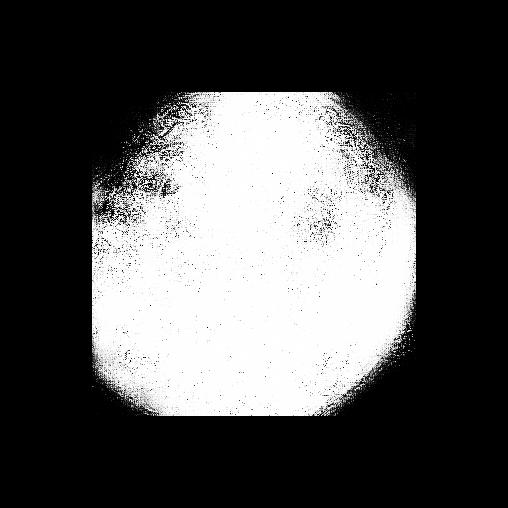} 
    \\
    & \rowname{BYOL} & 
	\includegraphics[width=0.15\textwidth]{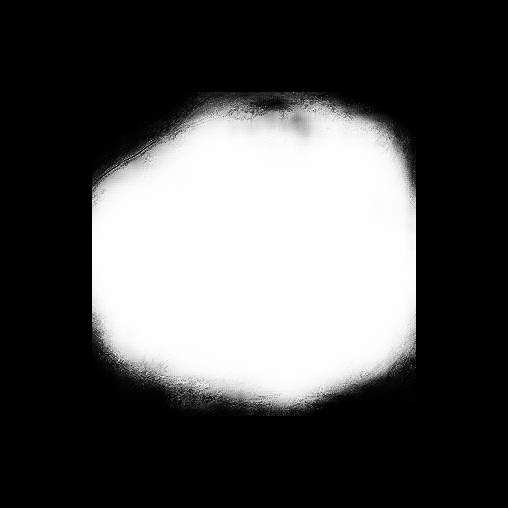} &
	\includegraphics[width=0.15\textwidth]{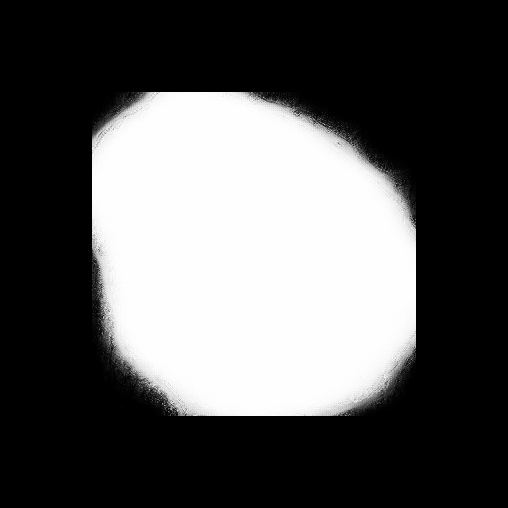} &
	\includegraphics[width=0.15\textwidth]{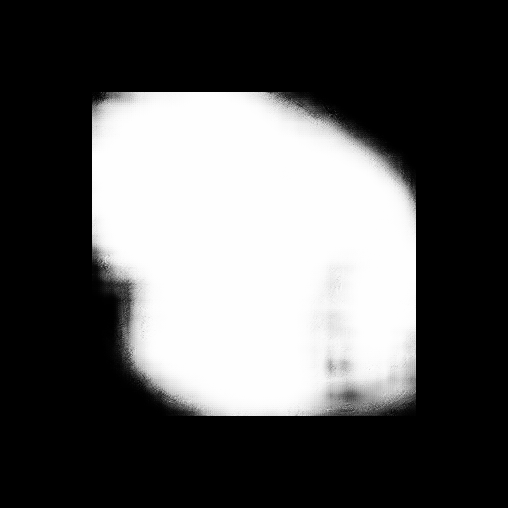} &
	\includegraphics[width=0.15\textwidth]{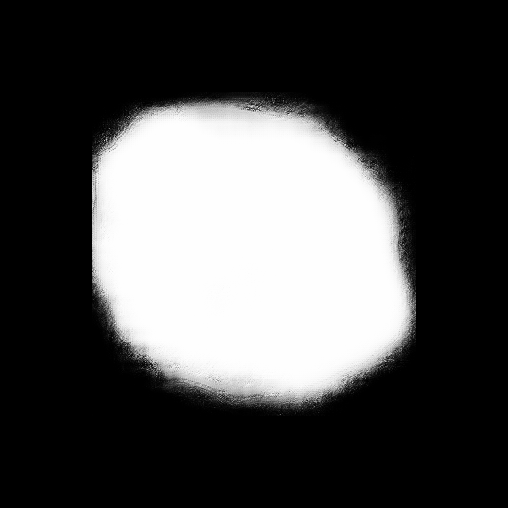} &
	\includegraphics[width=0.15\textwidth]{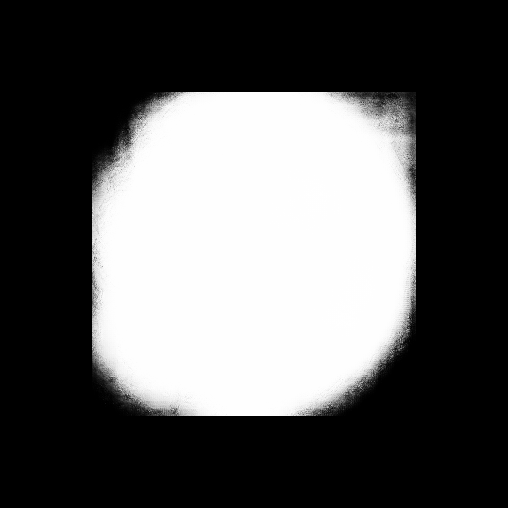} 

	\end{tabular}
	\caption{Visual comparison between predicted glomeruli segmentation maps and real ground-truths for each test stain using fine-tuned UNet and UDAGAN models (trained with 5\% labels).}
	\label{fig:qualitative_results}
\end{figure}

A visual confirmation of these findings is shown in Fig.\ \ref{fig:qualitative_results}, in which glomeruli segmentation maps (for models trained with 5\% labels) for each stain are presented.

\subsection{Omitting Validation Data} \label{appen_sec:without_validation_data}
As shown above, a balance between minimising labels and maximising performance is achieved using 5\% labels. Nevertheless, when training the final models, the results were obtained using a fully labelled validation set. Therefore, Table \ref{table:without_validation_set} evaluates whether the validation set is necessary or whether this labelling requirement can also be reduced. It is shown that in many cases, the performance without a validation set outperforms that obtained when using a labelled validation set. This is explained by the fact that in the dataset used, there is a lower domain shift (measured by following \cite{nisar2022towards}) between the train and test set distributions, which is $0.0655$ (averaged across all stains), compared to the train and validation set distributions, $0.1857$. This allows the models trained without validation data to outperform (on the test data) those selected using the validation loss. Although, this behaviour is specific to datasets with the above-mentioned characteristic, it only affects the difference in performance between the two experimental settings and not the findings themselves. Let us imagine that there were a lower domain shift between the validation and training sets, in this case removing the validation set would only eliminate the increase in performance observed here. It therefore does not invalidate the findings presented herein, that the validation set can be removed to further  minimise labelling requirements.

\begin{figure*}[tb] 
\centering \footnotesize
\begin{tabular}{c c} 
\includegraphics[width=0.5\textwidth]{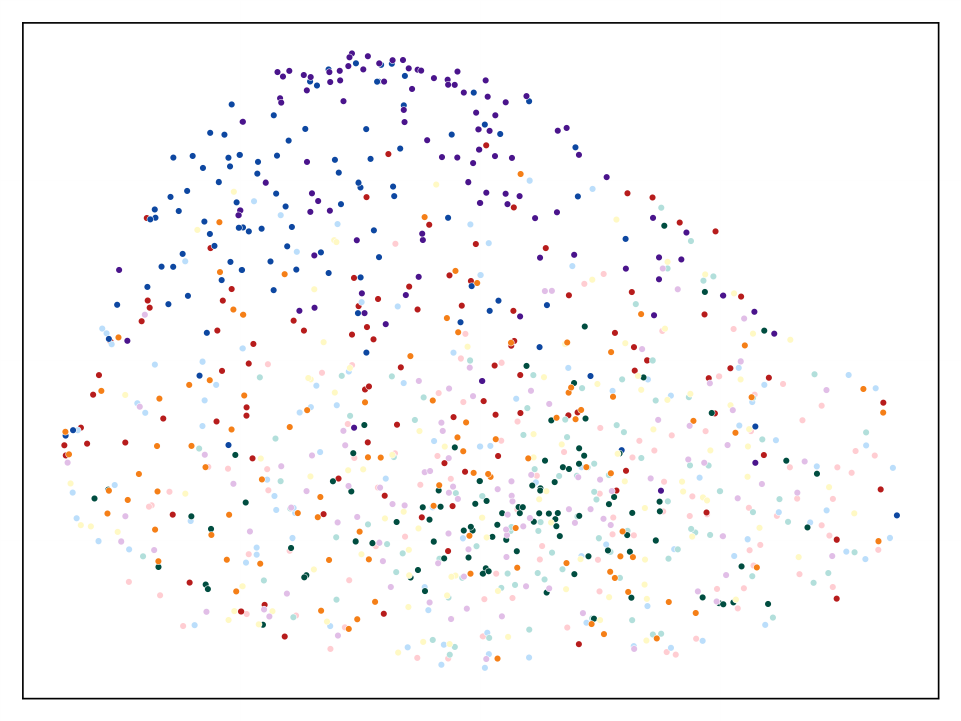} &
\includegraphics[width=0.5\textwidth]{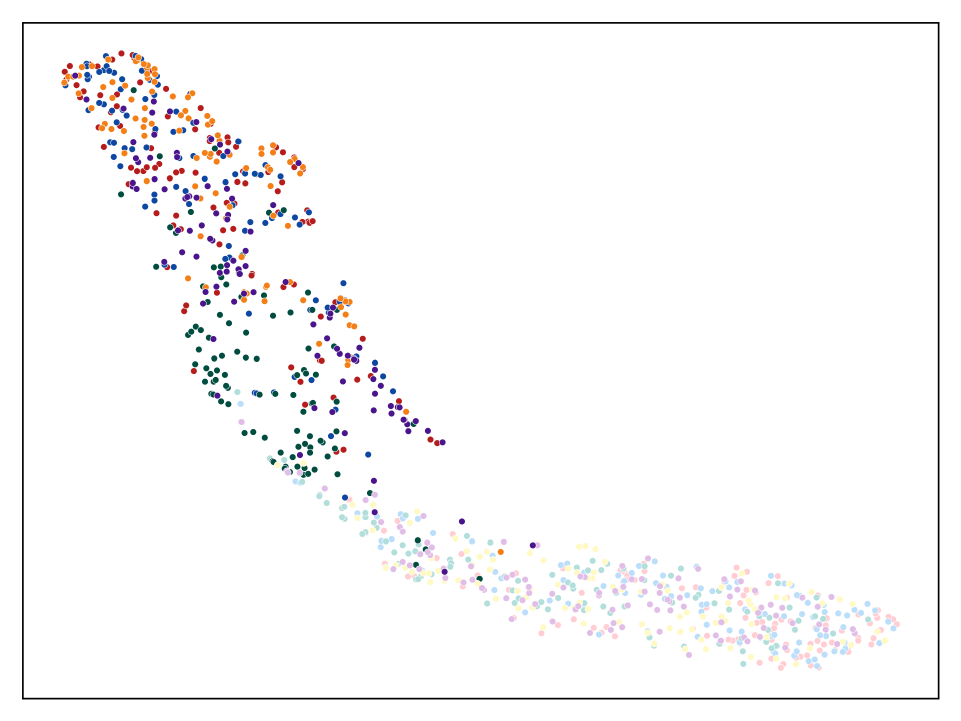} \\
(a) & (b) \\
\multicolumn{2}{c}{\includegraphics[width=0.5\textwidth]{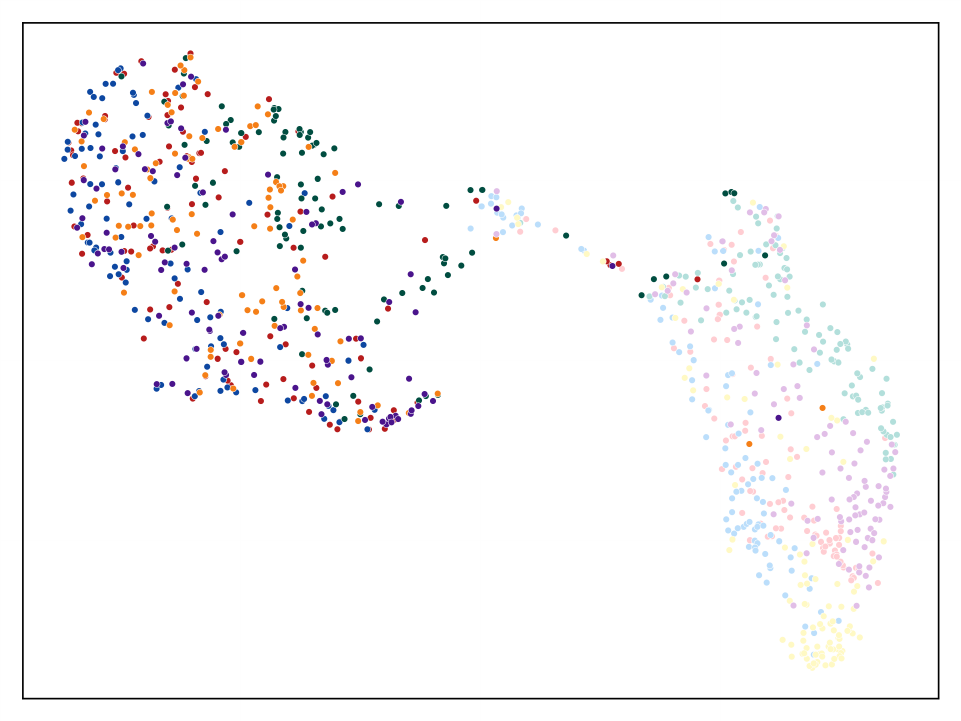}} \label{fig:check}\\ 
\multicolumn{2}{c}{(c)} \\
\multicolumn{2}{c}{\includegraphics[width=\textwidth]{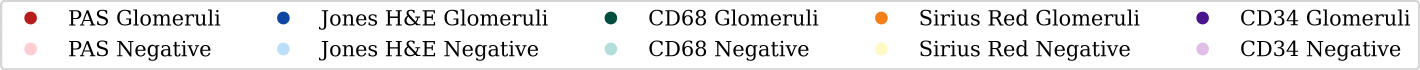}}
\end{tabular} 
\caption{Two-dimensional UMAP embeddings of the representations learned by: (a) SimCLR and (b) BYOL based UDAGAN models, trained without a validation set, and (c) SimCLR UDAGAN with a 5\% labeled validation set. Models randomly chosen, representations sampled from the penultimate convolutional layer, 100 patches per stain per class from the unseen test set. Each point is a patch from the respective class and staining.} 
\label{fig:umap_distribution_plots} 
\extralabel{fig:simclr_udagan_no_validation_set}{(a)}
\extralabel{fig:byol_udagan_no_validation_set}{(b)}
\extralabel{fig:simclr_udagan_with_validation_set}{(c)}
\end{figure*}

\begin{table*}[tb]
\caption{Downstream task performance with 5\% training labels, without a validation set. UNet, 5\% labels are used for all stains; UDAGAN, 5\% labels are used for only source, PAS, stain. The evaluation is conducted on test set. Each F$_{1}$ score is the average of five different training repetitions (standard deviations in parentheses). Highest F$_{1}$ score for each stain is in italics, overall highest F$_{1}$ score averaged across all stains is in bold.}
\label{table:without_validation_set}
\centering \scriptsize
\addtolength{\leftskip}{-1cm}
\begin{NiceTabular}{c:c:ccccc:c}
\toprule
\Block{2-1}{Downstream \\ Tasks} & \Block{2-1}{Pre-training} & \Block{1-5}{Test Stains} & & & & & \Block{2-1}{Average} \\
\cmidrule{3-7}
 & & PAS & Jones H\&E & CD68 & Sirius Red & CD34 & \\

\midrule
\Block{3-1}{UNet}& SimCLR & \textit{0.812 (0.019)} & 0.795 (0.034) & 0.575 (0.146) & 0.612 (0.066) & 0.810 (0.020) & 0.720 (0.057)\\ 
& BYOL & 0.786 (0.020) & \textit{0.839 (0.025)} & \textit{0.771 (0.027)} & \textit{0.788 (0.021)} & \textit{0.870 (0.003)} & \textbf{0.810 (0.019)}\\
& HR-CS-CO & 0.777 (0.032) & 0.695 (0.092) & 0.428 (0.086) & 0.425 (0.094) & 0.700 (0.060) & 0.605 (0.072)\\ 


\midrule
\Block{2-1}{UDAGAN} & SimCLR & 0.402 (0.193) & 0.389 (0.078) & 0.000 (0.000) & 0.072 (0.120) & 0.359 (0.260) & 0.244 (0.130)\\ 
& BYOL & \textit{0.850 (0.008)} & \textit{0.822 (0.021)} & \textit{0.650 (0.029)} & \textit{0.815 (0.026)} & \textit{0.771 (0.011)} & \textbf{0.765 (0.022)}\\

\bottomrule

\end{NiceTabular}
\end{table*}

With SimCLR and UDAGAN, however, a considerable drop in performance is observed. This is likely because of overfitting in the absence of validation data. The model is trained in two stages: (1) pre-training using SimCLR on original image patches; (2) translation (using CycleGAN models) from PAS to all other stains during fine-tuning. During the second stage, imperceptible noise caused by the CycleGAN transfer \cite{nisar2022towards} is introduced into the training patches. This causes a domain shift between the training data and test images, reducing performance. This is exacerbated by the absence of a validation set, which would normally prevent overfitting to this `noisy' training data. In contrast, BYOL is not affected because it uses batch-normalisation, which helps to stabilise the training process and prevent overfitting the noisy inputs. This can be visualised in Fig.\ \ref{fig:umap_distribution_plots}, where there is a noticeable lack of class boundary between test glomeruli and negative patches when training SimCLR-UDAGAN without a validation set, see Fig.\ \ref{fig:simclr_udagan_no_validation_set}. Such a boundary exists in the BYOL-UDAGAN representation without validation data, see Fig.\ \ref{fig:byol_udagan_no_validation_set}, and a SimCLR-UDAGAN trained with 5\% validation labels, see Fig.\ \ref{fig:simclr_udagan_with_validation_set} (for comparison, this model achieves an average F$_1$ score of $0.686$, vs.\ $0.244$ without the validation set).

\section{Generalisation and Reproducibility} \label{sec:generalisation}
To evaluate how well the pre-trained models---SimCLR, BYOL, and the proposed HR-CS-CO---generalise beyond our private dataset, we evaluate their performance on two widely recognised public datasets for kidney glomeruli segmentation. These include: (1) the ``HuBMAP --- Hacking the Kidney'' Kaggle Competition dataset \cite{jain2023segmentation}; and the ``Kidney Pathology Image Segmentation (KPIs)'' Challenge dataset \cite{deng2025kpis}. A short overview of these datasets is provided below:
\begin{description}[leftmargin=0pt]
\item[HuBMAP:] The HuBMAP dataset \cite{jain2023segmentation} is developed by the BIOmolecular Multimodal Imaging Center (BIOMIC) team at Vanderbilt University. It contains 30 WSIs stained with PAS and scanned at a high-resolution of 0.5 \si{\micro\meter} / pixel, producing images averaging around 33k$\times$31k pixels. The dataset is split into 15 WSIs for training containing 3770 glomeruli, 5 for validation (1388 glomeruli), and 10 for testing (2017 glomeruli). 

\item[KPIs:] The KPIS Challenge \cite{deng2025kpis} provides 50 high-resolution WSIs derived from four different groups of mouse models: (1) Normal - normal/healthy mice; (2) 56Nx - mice subjected to 5/6 nephrectomy; (3) DN - eNOS-/-/ lepr(db/db) double-knockout mice; and (4) NEP25 - transgenic mice that express human CD25 selectively in podocytes. All tissue samples are stained with PAS and captured at 40$\times$ digital magnification (a resolution of 0.110 \si{\micro\meter} / pixel), resulting in WSIs ranging from 14k$\times$16k to 81k$\times$70k pixels. Three expert pathologists annotated all the glomeruli using Qupath software \cite{bankhead2017qupath}. The whole dataset is divided into 30 WSIs for training, 8 for validation, and 12 for testing, containing 4443, 1374, and 1861 glomeruli, respectively.
\end{description}

\begin{figure}[!tb]
\centering \footnotesize
\settoheight{\tempdima}{\includegraphics[width=0.2\textwidth]{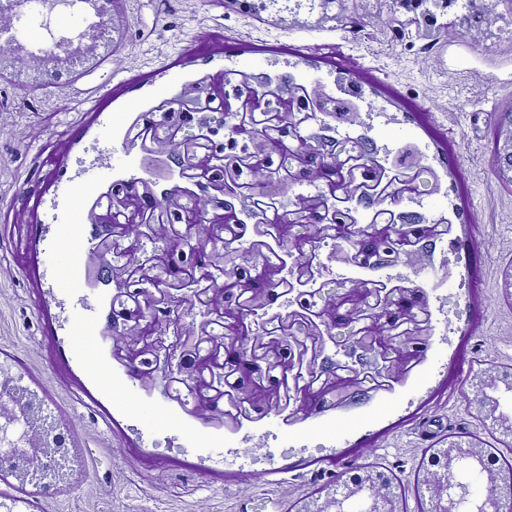}}
\begin{tabular}{c@{ }c@{ }c@{ }}
\includegraphics[width=0.2\textwidth]{generalisation/intra-stain_variability/IFTA_Nx_0011_02_glomeruli_patch_43.png} &
\includegraphics[width=0.2\textwidth]{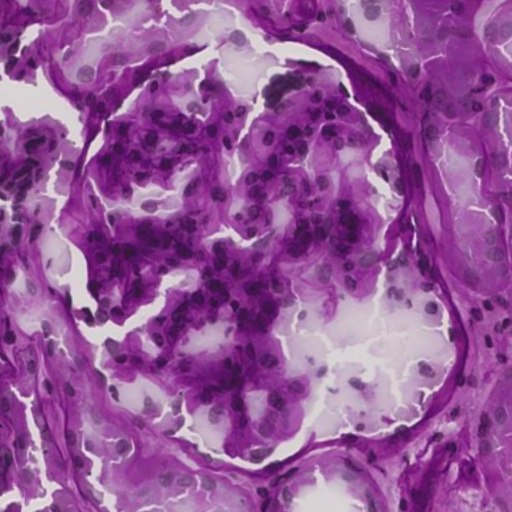} &
\includegraphics[width=0.2\textwidth]{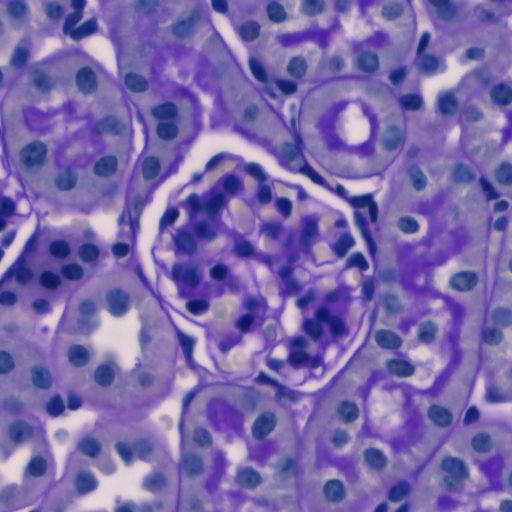} \\
(a) & (b) & (c)
\end{tabular}
\caption{Illustration of PAS-stained kidney tissue images from three datasets, with each image representing a glomerulus. Representative samples are shown from (a) our in-house dataset, (b) the HuBMAP dataset, and (c) the KPIs dataset.}
\label{fig:intra_stain_variability}
\end{figure}

While both datasets use the same PAS stain as included in our original pre-training set, there exists significant differences in the visual appearance of these stains (see Figure \ref{fig:intra_stain_variability}). These discrepancies stem from variations in tissue preparation methods, laboratory-specific techniques and the scanning devices \cite{vasiljevic2022cyclegan}. As a result, evaluating our pre-trained models on these datasets offers a rigorous evaluation of their generalisation capabilities beyond the original data. Since both HuBMAP and KPIs contain only single (PAS) stain samples, we used UNet as the downstream segmentation task, following the similar training protocols as detailed in our main UNet experiments (see Section \ref{sec:training::subsec:segmentation}). 

As found in the previous section, an effective balance between minimising annotation effort and maximising performance is achieved using 5\% labels. Therefore, in this study, all models were trained and evaluated using only 5\% labels. For comparison, we include results from fully supervised models (using 100\% labels). To provide broader context and benchmarking, the performance of top-performing methods \cite{singh2023glomerulus, andreini2024enhancing, liu2024hybrid, wajeeh2025leveraging} from recent literature (2023--2025) specifically designed and evaluated on these datasets is also reported. In addition, we report the performance of competition winning models such as UNet-SEResNext-101 \cite{jain2023segmentation}, which secured 1$^{\text{st}}$ place in the HuBMAP competition, and Mask2Former-Swin-B \cite{cap2025generalpipelineglomeruluswholeslide}, the top performer for the KPIs challenge.

\begin{table}[tb]
\caption{Comparison of generalisation performance between self-supervised pre-training methods (trained using 5\% labelled data) and their respective baselines (without pre-training) on publicly available kidney glomeruli segmentation datasets. For comparison, the results from fully supervised models (trained with 100\% labels) are included, as well as top-performing methods from literature and competition winning models. Performance is evaluated using mean F$_1$ score with standard deviations in parentheses. Best results for each Dataset/Label Split are shown in bold.}
\label{table:generalisation_results_five_percent}
\centering \scriptsize
\begin{NiceTabular}{ccccc}
\toprule
Dataset & Label & Method & Publication Year & F$_{1}$ Score \\
\midrule
\Block{13-1}{HuBMAP} & \Block{4-1}{5\%} & None (Baseline) & --- & 0.719 (0.005) \\
 & & SimCLR & --- & 0.736 (0.054) \\
 & & BYOL & --- & \textbf{0.816 (0.080)} \\
 & & HR-CS-CO & --- & \textbf{0.814 (0.052)} \\

 \cmidrule{2-5}
 & \Block{9-1}{100\%} &  None (Baseline) & --- & 0.947 (0.002) \\
 & & SimCLR & --- & 0.922 (0.025) \\
 & & BYOL & --- &0.913 (0.022) \\
 & & HR-CS-CO & --- & 0.920 (0.015) \\
 
 & & U-Net SEResNext-101$^\dagger$ \cite{jain2023segmentation} & 2023 & \textbf{0.951 (0.013)} \\
 & & LinkNet \cite{singh2023glomerulus} & 2023 & 0.942 (--) \\
 & & SegNeXt \cite{andreini2024enhancing} & 2024 & 0.868 (--) \\
 & & CNN-TransXNet \cite{liu2024hybrid} & 2024 & 0.828 (--) \\
 & & V-SAM-VIT-B \cite{wajeeh2025leveraging} & 2025 & \textbf{0.955 (--)} \\

\midrule
\Block{10-1}{KPIs} & \Block{4-1}{5\%} & None (Baseline) & --- & 0.672 (0.047) \\
 & & SimCLR & --- & 0.751 (0.008) \\
 & & BYOL & --- & \textbf{0.795 (0.016)} \\
 & & HR-CS-CO & --- & 0.651 (0.005) \\

 \cmidrule{2-5}
 & \Block{6-1}{100\%} &  None (Baseline) & --- & 0.786 (0.023) \\
 & & SimCLR & --- & 0.848 (0.016) \\
 & & BYOL & --- &0.837 (0.014) \\
 & & HR-CS-CO & --- &0.794 (0.013) \\

 & & HoloHisto-4K \cite{tang2024holohisto} & 2024 & 0.845 (--) \\
 & & Mask2Former-Swin-B$^\dagger$ \cite{cap2025generalpipelineglomeruluswholeslide} & 2025 & \textbf{0.886 (0.058)} \\

\bottomrule
\end{NiceTabular}
\caption*{$^\dagger$ indicates competition-winning method on the respective dataset (HuBMAP: U-Net SEResNext-101; KPIs: Mask2Former-Swin-B), reflecting the leaderboard’s top performance.}

\end{table}

Table \ref{table:generalisation_results_five_percent} presents the F$_1$ based segmentation scores achieved using pre-trained models (trained on our private dataset) and fine-tuned on HuBMAP and KPIs with both limited (5\%) and fully (100\%) labelled data. These results are closely aligned with performance trends observed in the previous section (on our private dataset), providing strong evidence on the generalisation capability of these pre-trained models when applied to new and unseen datasets.

The results show that the pre-trained models have consistent and often substantial gains compared to their baseline counterparts across both datasets. Notably, on the HuBMAP dataset, both BYOL and HR-CS-CO, achieve F$_1$ score of 0.816 and 0.814, respectively, when pre-trained on our dataset and fine-tuned using 5\% labels, representing a significant improvement of nearly 10\% over the baseline model (0.719). These performance gains are even more pronounced on the KPIs dataset, where BYOL achieves the highest F$_1$ score of 0.795 with just 5\% labels. This not only reflects a 12.3\% increase over the respective baseline (0.672), but also exceeds the performance of fully supervised baseline (0.786) trained with 100\% labels. These findings highlight the effectiveness of these pre-trained models in extracting transferable visual features that facilitate robust performance even with heavily  reduced labels. The only notable exception is HR-CS-CO on the KPIs dataset, where its performance (0.651) with 5\% labels is slightly below the baseline.

Beyond surpassing the baseline models, the pre-trained models demonstrate a substantial ability to close the performance gap with fully supervised and state-of-the-art benchmarks. For instance, on the HuBMAP dataset, the top performing pre-trained models---BYOL and HR-CS-CO---trained with only 5\% labels, achieve over 85\% of the F$_1$ score of the competition winning fully supervised method (U-Net SEResNext-101, F$_1$=0.951). These models also show competitive performance relative to other leading fully supervised methods such as LinkNet and SegNeXt. On KPIs, BYOL again stands out with an F$_1$ score of 0.795. This not only surpasses the fully supervised baseline (0.786) but also significantly reduces the performance gap with other fully supervised alternatives (HoloHisto-4K, F$_1$=0.845 and Mask2Former-Swin-B, F$_1$=0.886). Moreover, when trained with 100\% labelled data, all pre-trained models (except HR-CS-CO on KPIs) outperform the baseline model with a margin of 5--6\%, and their performance closely approaches that of theg top-ranking model (Mask2Former-Swin-B). These findings highlight the effectiveness of these pre-trained models in learning robust and transferable features that enable competitive performance even under extreme annotation scarcity (95\% reduced annotations).

To facilitate reproducibility and further research, all our pre-trained models, as well as fine-tuning scripts and detailed instructions, have been made publicly available\footnote{\url{https://github.com/zeeshannisar/maximising-kidney-glomeruli-segmentation-using-limited-labels}}. 

\section{Discussion} \label{sec:discussions}
The previous sections have demonstrated the effectiveness of SSL in combating a lack of labelled segmentation data in histopathology, with pre-trained models approaching fully-supervised performance (e.g.\ with BYOL) in both single-stain UNet and multi-stain UDAGAN models (e.g.\ with $\sim$30 annotated images per training stain). 

We can observe however that not all self-supervised learning approaches are equal. When fewer labels are available (1\% and 5\%), general computer vision (CV) approaches such as SimCLR and BYOL perform best. Even though HR-CS-CO is specifically designed for histopathology, it only becomes competitive and/or outperforms the CV approaches when provided with moderate (10\%) to larger amounts of labelled data. It is particularly successful when applied to the CD68 stain, outperforming even the baseline models. CD68 is an immunohistochemical stain in which haematoxlyn highlights the main structural component and specific immune cells are highlighted in brown. It is therefore particularly suited to an approach such as HR-CS-CO. There are many other similar immunohistochemical stainings, and more complicated double stainings (e.g.\ CD3-CD68, CD3-CD163, CD3-CD206, etc \cite{MERVEILLE2021106157}) that should be suitable for such an approach (including the H\&E stain CS-CO was originally developed for). In stains such as Sirius Red, 
it appears that the superposition of staining components (and weak haematoxyln staining) prevents the haematoxlyn channel from being efficiently extracted, limiting the effectiveness of HR-CS-CO (the difficulty of extracting this component from Sirius Red has been previously noted in the literature \cite{lampert2019strategies}). 

The intended application of the pre-trained model naturally dictates the type of SSL that should be used. In UDAGAN, BYOL consistently outperforms SimCLR, especially with highly limited labels, likely due to its robustness to noise during fine-tuning. Unlike SimCLR, which relies on negative pairs, BYOL uses only positive samples and keeps a moving average for regularisation, making it less sensitive to noisy (i.e.\ translated) data during fine-tuning \cite{lee2021compressive}.  
It is known that this noise is particularly evident in immunohistochemical stainings such as CD68 and CD34 \cite{vasiljevic2021towards,nisar2022towards}, which is confirmed in this study where the noise degrades the performance of all pre-training methods equally. 

The role of validation data was shown to strongly impact the success of fine-tuning pre-trained models. Surprisingly, omitting a validation set greatly improved the success of fine-tuning, reaching performance levels approaching those of fully supervised models. This means that almost state-of-the-art performance can be achieved while reducing labelling requirements by 95\%.

Finally, the benefits of self-supervised pre-training are not just restricted to limited label situations. This study has shown that the performance of fully-supervised stain-invariant models such as UDAGAN can be improved, meaning that pre-training the UDAGAN model before fully-supervised training leads to a 2.6\% increase in F$_1$ score. This offers a new SOTA performance in stain-invariant glomeruli segmentation without any architectural nor labelling changes. 

Moving away from renal histopathology, these results are consistent with other histopathology studies found in the literature and extend upon existing efforts to reduce the need for extensive manual annotations. Particularly, \citet{prakash2020leveraging} showed that for nuclei segmentation in the Broad Bioimage Benchmark Collection dataset, a self-supervised fine-tuned UNet using only 5\% labels (32 images) demonstrated only 3\% reduction in IoU score compared to a full supervised UNet.
Similarly, \citet{punn2022bt} reported that a self-supervised fine-tuned UNet using 20\% labels (134 images) for nuclei segmentation on the Kaggle Datascience Bowl Challenge 2018 dataset only lost 5.1\% in F$_1$ score compared to a fully supervised UNet.
Combined with the results presented herein, these demonstrate  minimal performance degradation despite significant reductions in label requirements. The findings presented herein, however, go further. Not only do they show the benefit of integrating pre-training into fully-supervised approaches, but also into multi-stain segmentation strategies and remove the need for labelled validation datasets. This leads to a significant decrease in  labelling requirement to the source stain (a reduction of at least $n$ times, where $n$ is the number of stains to be segmented).

This discussion has already outlined the limitations of HR-CS-CO and so it remains to address SSL limitations in general. Foremost, there is a risk of introducing false negatives when training SimCLR on datasets with few classes because mini-batches are likely to contain several samples from one class. This can lead to a model that fails to distinguish between semantically ``similar'' and ``dissimilar'' images, reducing downstream performance. BYOL, however, overcomes this limitation by not using negative pairs. Moreover, contrastive SSL in general relies on augmentation to create ``similar'' pairs. As outlined by \citet{garcea2022data}, medical imaging is sensitive to augmentation since it contains subtle, easily distorted features. 

\section{Conclusions} 
\label{sec:conclusion}
This article has shown how to significantly reduce  the need for labelled data ($>$95\%) in histopathology image segmentation. To achieve this, self-supervised pre-training techniques---SimCLR, BYOL, and a novel histopathological SSL approach, HR-CS-CO---were used to learn general features from unlabelled data. These features were then fine-tuned for single stain and multi-stain segmentation tasks using UNet, and UDAGAN models, making them robust to training scenarios with limited labels. 

These approaches demonstrated consistently superior performance compared to their respective baselines, and were able to approach the performance of fully supervised models. Moreover, the pre-trained models were generalisable beyond their training datasets. These findings underscore the potential and significance of incorporating these advanced learning techniques in histopathology. The results also demonstrated that self-supervised learning combined with fine-tuning is most effective without a validation set, further reducing the labelling requirement. However, some methods, such as SimCLR, are more susceptible to domain shifts and may benefit from some labelled validation data to ensure generalisation.

Furthermore, this study advanced the recent trend in histopathology towards creating multi-stain segmentation models by demonstrating that it is possible to train a stain-invariant segmentation model with as few as 5\% labels from only one stain. This model closely matches the performance of a fully supervised UNet trained with $N \times 100\%$ labels, where $N$ is the number of stains.

\section*{CRediT authorship contribution statement}
\noindent \textbf{ZN:} Conceptualisation, Methodology, Software, Validation, Investigation, Visualisation, Writing - original draft. \textbf{FF:} Data Curation, Validation. \textbf{TL:} Conceptualisation, Validation, Supervision, Funding acquisition, Writing - review \& editing. 


\section*{Acknowledgements}
Funded by ANR HistoGraph (ANR-23-CE45-0038) and the ArtIC project ``Artificial Intelligence for Care'' (grant ANR-20-THIA-0006-01), co funded by \textit{Région Grand Est}, Inria Nancy - Grand Est, IHU Strasbourg, University of Strasbourg \& University of Haute-Alsace. We acknowledge the ERACoSysMed \& e:Med initiatives by BMBF, SysMIFTA (managed by PTJ, FKZ 031L-0085A; ANR, grant ANR-15-CMED-0004), Prof.\ C\'{e}dric Wemmert, and Prof.\ Friedrich Feuerhake and team at MHH for the high-quality images \& annotations: 
N.\ Kroenke,
N.\ Schaadt,
V.\ Volk
\& J.\ Schmitz.
We thank Nvidia, the \textit{Centre de Calcul} (University of Strasbourg) \& GENCI-IDRIS (grant 2020-A0091011872) for GPU access. 





\bibliographystyle{elsarticle-num-names} 
\bibliography{main.bib}
\pagebreak

\appendix






\section{Methods} \label{appen_sec:methods}
\subsection{Self-supervised Pre-training}
\subsubsection{SimCLR} \label{appendix:subsubsec:simclr}
\begin{figure}[tb]
    \centering
    \includegraphics[width=0.8\textwidth]{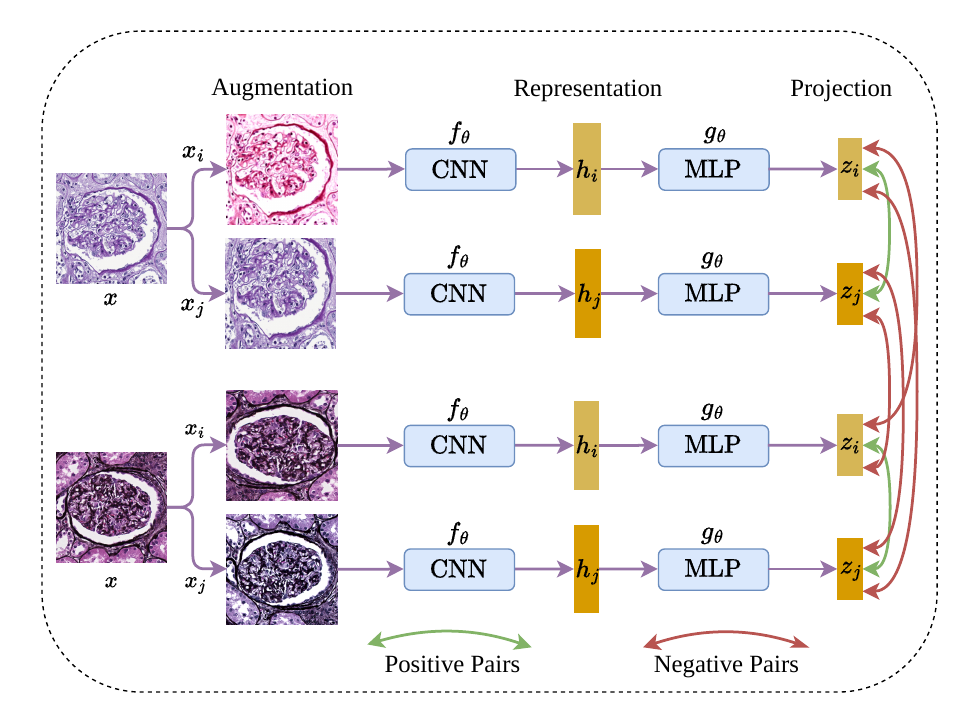}
    \caption{Overview of the SimCLR architecture inspired by \cite{chen2020simple, ciga2022self}.}
    \label{fig:simclr}
\end{figure}
SimCLR \cite{chen2020simple} learns representations by maximising the agreement between two augmented views of the same image via contrastive loss in the latent space. As shown in Fig.\ \ref{fig:simclr}, the framework starts with a probabilistic data augmentation module $f_{aug}$ that generates two positively correlated views, $x_i$ and $x_j$, of a given data sample $x$. A set of base augmentations \cite{chen2020simple} are adopted (using the Albumentations library \cite{buslaev2020albumentations} and parameters from \cite{stacke2021learning}), including random cropping and resizing with a large scale range of ($0.2-1.0$), flipping, grey-scale, Gaussian blur, and random colour distortions. Based on the findings of \citet{stacke2021learning}, two additional augmentations were used, grid distort and grid shuffle, which have demonstrated their effectiveness for histopathology related applications. 
The augmented views, $x_i$ and $x_j$ are then transformed into their corresponding representations, $h_i$ and $h_j$ by employing a convolutional neural network (CNN) base encoder $f_{\theta}$, where $\theta$ is the weight parameters.


Subsequently, a projection head $g_{\theta}$ consisting of a multi-layer perceptron (MLP) is employed to map the extracted representations into a lower embedding space in which the contrastive loss is applied. The MLP comprises two dense layers with ReLU activation for the first layer; and linear activation for the second layer to obtain $z_i = g_{\theta}(h_i)$ and $z_j = g_{\theta}(h_j)$ respectively. In \cite{chen2020simple}, it was observed that comparing $z_i$ and $z_j$ was more effective for learning representations than directly comparing $h_i$ and $h_j$. Finally, as suggested by the authors of SimCLR, to optimise the entire network the NT-Xent (the normalised temperature-scaled cross-entropy loss) function is defined, such that
\begin{equation}
\ell_{i, j} = -\log\frac{\exp\left(sim(z_i, z_j)/\tau\right)}{\sum_{k=1}^{2N}\mathds{1}_{[k \neq i]}\exp\left(sim(z_i, z_{k})/\tau\right)},
\label{eq:simclr}
\end{equation}
where $\tau$ is the temperature parameter that weights different samples and facilitates learning from hard negative samples and $\mathds{1}$ is the indicator function. The term $sim(z_i, z_j) = z_{i}^\top z_j/\|z_i\| \|z_j\|$ represents the dot product between $\ell_{2}$ normalised $z_i$ and $z_j$, which corresponds to the  cosine similarity. This loss functions aims to maximise the agreement between positive pairs of augmented images, while minimising it for other images in the same batch (negative pairs). In each training step with a batchsize of $2N$, each augmented image has one positive and $2(N-1)$ negative pairs.

\subsubsection{BYOL} \label{appendix:subsubsec:byol}
\begin{figure}[tb]
    \centering
    \includegraphics[width=\textwidth]{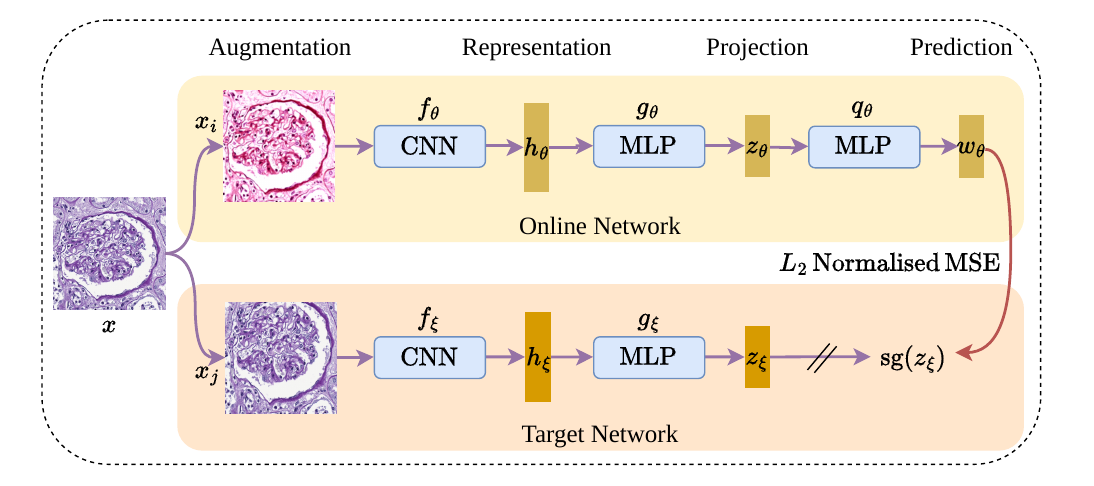}
    \caption{Overview of BYOL architecture inspired by \citet{grill2020bootstrap}.}
    \label{fig:byol}
\end{figure}
Bootstrap Your Own Latent Representation (BYOL) is an implicit contrastive learning approach introduced by \citet{grill2020bootstrap}. It does not rely on negative pairs and is more robust to the choice of augmentations, resulting in superior performance compared to other contrastive methods. The core idea of BYOL is to iteratively bootstrap the network's output to serve as a target for an enhanced representation. To achieve this, BYOL employs two neural networks, Online and Target, which interact and learn from each other.  As depicted in Fig.\ \ref{fig:byol}, the Online network is a trainable network comprising a CNN based encoder $f_{\theta}$, an MLP based projection head $g_{\theta}$, and a prediction head $q_{\theta}$. On the other hand, the Target network is a non-trainable network that is randomly initialised. It has the same architecture as the Online network, but has a different set of weight parameters $\xi$. The Target network provides the regression targets, used to train the Online network, and its parameters $\xi$ are updated through an exponential moving average of the Online parameters $\theta$. Considering a target decay rate $\tau \in [0, 1]$, the following update is carried out after each training step:
\begin{equation}
    \xi \leftarrow \tau \xi + (1 - \tau)\theta.
\end{equation}
To train the BYOL network, a data augmentation module $f_{aug}$ is used to generate two distinct augmented views $x_{i}$ and $x_{j}$ from the input image $x$. This module incorporates the same augmentations as described for SimCLR.
The Online network processes the first augmented view $x_{i}$ and outputs a representation $h_{\theta}$, a projection $z_{\theta}$, and a prediction $w_{\theta}$. Similarly, the Target network outputs a representation $h_{\xi}$, and a target projection $z_{\xi}$ from the second augmented view $x_{j}$. Notably, the prediction head is solely applied to the Online network, resulting in an asymmetric architecture  between the Online and Target pipelines. Following that, both $w_{\theta}$ and $z_{\xi}$ are normalised using $\ell_{2}$ norm and then fed into a mean squared error (MSE) loss function, such that
\begin{equation}
    \mathcal{L}_{\theta, \xi} = \left\| \bar{w}_{\theta} - \bar{z}_{\xi}\right\|^2_2 = 2 - 2 \cdot \frac{\langle w_{\theta}, z_{\xi} \rangle}{\left\|w_{\theta}\right\|_{2} \cdot \left\|z_{\xi}\right\|_{2}}.
\end{equation}
The loss $\mathcal{L}_{\theta, \xi}$ is made symmetrical by separately feeding $x_{j}$ to the Online network and $x_{i}$ to the Target network. This allows the computation of another loss function $\tilde{\mathcal{L}}_{\theta, \xi}$. During each training step, a stochastic optimisation step is performed to minimise $\mathcal{L}^{BYOL}_{\theta, \xi} = \mathcal{L}_{\theta, \xi} + \tilde{\mathcal{L}}_{\theta, \xi}$ with respect to $\theta$ only, while $\xi$ remains unaffected by applying a stop-gradient ($sg$) as shown in Fig.\ \ref{fig:byol}.

\section{Supplementary Results}
\subsection{Fixed Representations} \label{appen_sec:fixed_features_results}
To evaluate whether self-supervised learning methods are able to learn meaningful representations and generalise to downstream tasks, it is important to use a fixed-feature setting. Therefore, in this setting, the representations learned are used as feature vectors in the downstream tasks of UNet, MDS1, and UDAGAN and the results are provided in Table \ref{table:fixed_features_results}, evaluated across different splits of labelled data. Since these self-supervised pre-training methods are not explicitly designed for learning representations well-suited for the tasks of UNet, MDS1, and UDAGAN. Consequently, fixed-feature settings exhibit a significant drop in performance when compared to fine-tuned models (as presented in Table \ref{table:finetune_results}), and this is particularly noticeable in the case of HR-CS-CO. This highlights the need for a more effective stain separation methods beyond the classical, matrix decorrelation based approach employed in our study. This is why, during  fine-tuning,  HR-CS-CO's representation is able to better adapt to the specific characteristics of the downstream task, and therefore compensate for the limitations caused by the loss of information resulting from the stain separation used during training. Nonetheless, it is essential to acknowledge that even though fixed-feature models experience a decline in performance, they show improved results in comparison to baseline models, especially when employing SimCLR and BYOL as pre-trained models. This improvement is particularly evident when the models are subjected to limited labelled data, such as 1\% and 5\%. Moreover, when provided with moderate (10\%) to fully (100\%) labelled data, the fixed-feature models approach the performance levels of baseline models. This highlights the effectiveness of self-supervised learning methods in the context of their capacity to learn meaningful representations. 
\begin{table*}[!tb]
\centering \scriptsize
\addtolength{\leftskip} {-1.5cm}
\addtolength{\rightskip}{-1.5cm}
\caption{Performance evaluation of various self-supervised learning based UNet methods in a fixed-feature scenario for glomeruli segmentation. The performance is evaluated in terms of segmentation (F$_1$) score, averaged over five different training repetitions, with the standard deviations presented in parentheses.}
\label{table:fixed_features_results}
\begin{NiceTabular}{c:c:c:ccccc:c}
\toprule
\Block{2-1}{Downstream\\Tasks} & \Block{2-1}{Label\\Splits} & \Block{2-1}{Pre-training} & \Block{1-5}{Test Stains} & & & & & \Block{2-1}{Average}\\
\cmidrule{4-8}
& & & PAS & Jones H\&E & CD68 & Sirius Red & CD34 & \\
\midrule
\Block{12-1}{UNet} 
 & \Block{3-1}{1\%} & SimCLR & 0.575 (0.043) & 0.472 (0.022) & 0.348 (0.073) & 0.376 (0.070) & 0.700 (0.032) & 0.494 (0.048)\\
 &  & BYOL & 0.478 (0.086) & 0.556 (0.075) & 0.190 (0.075) & 0.471 (0.071) & 0.688 (0.011) & 0.477 (0.064)\\
 &  & HR-CS-CO & 0.191 (0.049) & 0.079 (0.039) & 0.030 (0.061) & 0.170 (0.070) & 0.312 (0.075) & 0.156 (0.059)\\
\cmidrule{2-9}
 & \Block{3-1}{5\%} & SimCLR & 0.800 (0.009) & 0.724 (0.013) & 0.538 (0.087) & 0.526 (0.093) & 0.809 (0.004) & 0.679 (0.041)\\
 &  & BYOL & 0.734 (0.068) & 0.763 (0.035) & 0.435 (0.044) & 0.656 (0.047) & 0.745 (0.040) & 0.667 (0.047)\\
 &  & HR-CS-CO & 0.469 (0.056) & 0.546 (0.020) & 0.228 (0.051) & 0.252 (0.054) & 0.499 (0.036) & 0.399 (0.043)\\
\cmidrule{2-9}
 & \Block{3-1}{10\%} & SimCLR & 0.850 (0.005) & 0.794 (0.017) & 0.698 (0.033) & 0.509 (0.077) & 0.820 (0.019) & 0.734 (0.030) \\
 &  & BYOL & 0.785 (0.042) & 0.765 (0.015) & 0.600 (0.042) & 0.731 (0.036) & 0.789 (0.032) & 0.734 (0.033)\\
 &  & HR-CS-CO & 0.563 (0.051) & 0.644 (0.011) & 0.279 (0.016) & 0.461 (0.083) & 0.540 (0.034) & 0.498 (0.039)\\
\cmidrule{2-9}
 & \Block{3-1}{100\%} & SimCLR & 0.881 (0.006) & 0.848 (0.016) & 0.794 (0.011) & 0.786 (0.024) & 0.876 (0.009) & 0.837 (0.013)\\
 &  & BYOL & 0.878 (0.007) & 0.849 (0.011) & 0.781 (0.012) & 0.800 (0.018) & 0.867 (0.011) & 0.835 (0.012)\\
 &  & HR-CS-CO & 0.578 (0.077) & 0.711 (0.009) & 0.619 (0.032) & 0.683 (0.028) & 0.675 (0.013) & 0.653 (0.032)\\

\midrule

\Block{8-1}{UDAGAN} & \Block{2-1}{1\%} & SimCLR & 0.529 (0.038) & 0.463 (0.055) & 0.315 (0.078) & 0.491 (0.048) & 0.589 (0.043) & 0.477 (0.053)\\
& & BYOL & 0.534 (0.020) & 0.427 (0.018) & 0.281 (0.042) & 0.473 (0.051) & 0.560 (0.021) & 0.455 (0.031)\\ 

\cmidrule{2-9}
& \Block{2-1}{5\%} & SimCLR & 0.752 (0.007) & 0.664 (0.042) & 0.524 (0.067) & 0.689 (0.008) & 0.753 (0.010) & 0.677 (0.027)\\
& & BYOL & 0.779 (0.023) & 0.683 (0.043) & 0.462 (0.047) & 0.694 (0.026) & 0.701 (0.044) & 0.664 (0.037)\\ 

\cmidrule{2-9}
& \Block{2-1}{10\%} & SimCLR & 0.775 (0.019) & 0.691 (0.045) & 0.608 (0.035) & 0.733 (0.026) & 0.768 (0.011) & 0.715 (0.027)\\ 
& & BYOL & 0.830 (0.027) & 0.743 (0.029) & 0.518 (0.035) & 0.728 (0.031) & 0.764 (0.017) & 0.717 (0.028)\\ 

\cmidrule{2-9}


& \Block{2-1}{100\%} & SimCLR & 0.835 (0.018) & 0.755 (0.036) & 0.637 (0.056) & 0.794 (0.031) & 0.772 (0.032) & 0.758 (0.035)\\ 
& & BYOL & 0.860 (0.020) & 0.819 (0.022) & 0.618 (0.025) & 0.810 (0.022) & 0.791 (0.025) & 0.780 (0.023)\\ 

\bottomrule
\end{NiceTabular}
\end{table*}


%
\end{document}